\pdfoutput=1
\documentclass[11pt]{article}

\usepackage[table]{xcolor}
\usepackage{amsmath}
\usepackage{xurl}
\usepackage{booktabs}
\usepackage{array,multirow}
\usepackage{graphicx}        
\usepackage{url}
\usepackage{amssymb}
\usepackage{amsfonts}
\usepackage{caption}
\usepackage{subcaption}
\usepackage{here}
\usepackage{tabularx}
\usepackage{comment}
\usepackage{xcolor}
\usepackage{enumitem}
\usepackage{hhline}
\usepackage{CJKutf8}
\usepackage{arabtex}
\usepackage{utf8}
\setcode{utf8}

\usepackage[]{acl}
\usepackage{times}
\usepackage{latexsym}
\usepackage[T1]{fontenc}
\usepackage[utf8]{inputenc}
\usepackage{microtype}
\usepackage{inconsolata}

\usepackage[most]{tcolorbox}
\newtcolorbox{recobox}[1][]{
  colback=white,    %
  colframe=black,   %
  fonttitle=\bfseries,     %
  title=Recommendation,    %
  #1                       %
}

\usepackage{fix-cm}

\makeatletter
\newcommand\notsotiny{\@setfontsize\notsotiny{7}{8}}
\makeatother

\title{An Empirical Study on Cross-lingual Vocabulary Adaptation \\for Efficient Language Model Inference}

\author{
Atsuki Yamaguchi$^{1}${\rm ,} Aline Villavicencio$^{1,2,3}$ \and Nikolaos Aletras$^{1}$\\ 
$^{1}$School of Computer Science, University of Sheffield, United Kingdom\\
$^{2}$Department of Computer Science, Institute of Data Science and Artificial Intelligence,\\
University of Exeter, United Kingdom\\
$^{3}$The Alan Turing Institute, United Kingdom\\
\texttt{\{ayamaguchi1,a.villavicencio,n.aletras\}@sheffield.ac.uk} 
}

\begin{document}
\maketitle

\begin{abstract}

The development of state-of-the-art generative large language models (LLMs) disproportionately relies on English-centric tokenizers, vocabulary and pre-training data.
Despite the fact that some LLMs have multilingual capabilities, recent studies have shown that their inference efficiency deteriorates when generating text in languages other than English. This results in increased inference time and costs.
Cross-lingual vocabulary adaptation (CVA) methods have been proposed for adapting models to a target language aiming to improve downstream performance. However, the effectiveness of these methods on increasing inference efficiency of generative LLMs has yet to be explored.
In this paper, we perform an empirical study of five CVA methods on four generative LLMs (including monolingual and multilingual models) across four typologically-diverse languages and four natural language understanding tasks.
We find that CVA substantially contributes to LLM inference speedups of up to 271.5\%. We also show that adapting LLMs that have been pre-trained on more balanced multilingual data results in downstream performance comparable to the original models.\footnote{Our code and models are available on  \href{https://github.com/gucci-j/llm-cva}{GitHub}.}
\end{abstract}

\section{Introduction} \label{sec:intro}
Generative large language models (LLMs) obtain strong generalization performance in many downstream natural language processing tasks~\cite{Achiam2023GPT4TR,Touvron2023LLaMAOA,Jiang2023Mistral7} across various languages. For example, BLOOM~\cite{Scao2022BLOOMA1} supports 46 languages while Open AI's ChatGPT reportedly supports 90 languages~\cite{ahuja-etal-2023-mega}.

\begin{figure}[!t]
\begin{center}
\includegraphics[width=0.8\columnwidth]{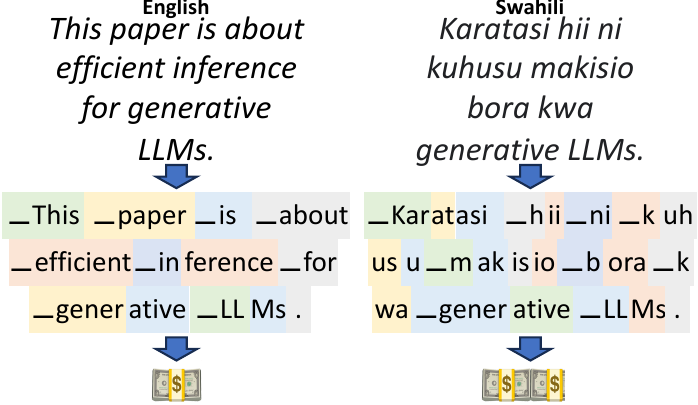}
\caption{
Example of overfragmentation when applying the Mistral-7B tokenizer to non-English text. 
}
\label{fig:motivation}
\end{center}
\end{figure}

Despite the multilingual capabilities of state-of-the-art LLMs, their development disproportionately relies on English-oriented tokenizers, vocabulary and pre-training data. For example, around 30\% of the training data in BLOOM is English.
This negatively affects the efficiency and downstream performance of LLMs in other languages. 
It has been demonstrated that LLMs overfragment text in underrepresented languages with different writing systems~\cite{rust-etal-2021-good,muller-etal-2021-unseen}, resulting to increased processing time, latency and costs for non-English speakers \citep{Ahia2023DoAL, petrov2023language}. Moreover, recent studies \cite{lin-etal-2022-shot,ahuja-etal-2023-mega,muennighoff-etal-2023-crosslingual} found that LLMs often perform better in a given language other than English when prompted in English instead of prompting directly in the other language. This is an unrealistic setting for non-English speakers that introduces extra disadvantages. Figure~\ref{fig:motivation} shows an illustrative example of overfragmentation in non-English text generation.

Cross-lingual vocabulary adaptation (CVA) is an efficient method for cross-lingual transfer~\cite{Tran2019FromET,wang-etal-2020-extending,chau-etal-2020-parsing}. The vocabulary of a source model is first updated (or replaced) with tokens from a target language, followed by fine-tuning the embedding matrix on data from the target language. Previous work on CVA primarily aims to improve downstream performance such as natural language inference and named-entity recognition~\cite{minixhofer-etal-2022-wechsel,Dobler2023FOCUSEE}. However, the effectiveness of these methods on improving inference efficiency of generative LLMs has yet to be explored.
We hypothesize that LLM inference in a target language can be improved by adapting the vocabulary of the source model to reduce text overfragmentation.

To test our hypothesis, we perform an empirical study of five CVA methods, on four generative LLMs, including a wide range of downstream tasks, from text classification, and span prediction, to summarization in zero-shot and few-shot settings across four diverse languages (i.e. German, Japanese, Arabic, and Swahili). Our contributions are as follows:
\begin{itemize}%
\item We demonstrate that CVA accelerates inference by up to 271.5\% in 99\% of cases (\S\ref{sec:speedup}).
\item We show that multilingual LLM vocabulary adaptation leads to comparable downstream performance to multilingual source LLMs (\S\ref{subsec:performance}).
\item We conduct an analysis to shed light on different design choices regarding the practical application of CVA in generative LLMs (\S\ref{sec:analysis}), and provide specific recommendations on how to select an optimal vocabulary initialization method following our analysis (\S\ref{sec:recommendation}).
\end{itemize} 

\section{Related Work} \label{sec:related}
\paragraph{Impact of Tokenization on LLMs}
Subword tokenization splits text into subword units and is 
the standard approach for tokenization in LLMs~\cite{Scao2022BLOOMA1,Touvron2023LLaMAOA,Jiang2023Mistral7}. It includes methods such as WordPiece~\cite{6289079}, Byte Pair Encoding (BPE)~\cite{sennrich-etal-2016-neural}, and Unigram~\cite{kudo-2018-subword}. Other approaches include word-~\cite{NIPS2000_728f206c,NIPS2013_9aa42b31}, character-~\cite{Al-Rfou_Choe_Constant_Guo_Jones_2019} and byte-level~\cite{xue-etal-2022-byt5} tokenization.

The impact of tokenization on LLMs has been actively studied including model performance~\cite{bostrom-durrett-2020-byte,rust-etal-2021-good,gow-smith-etal-2022-improving,10.1145/3578707,fujii-etal-2023-different}, inference speed~\cite{hofmann-etal-2022-embarrassingly,sun-etal-2023-multi,petrov2023language}, memory usage~\cite{sun-etal-2023-multi}, training~\cite{Ali2023TokenizerCF} and API costs~\cite{Ahia2023DoAL,petrov2023language}.
It is acknowledged that tokenizers lead to disproportionate fragmentation for different languages and scripts in multi- and cross-lingual settings~\cite{rust-etal-2021-good,muller-etal-2021-unseen}.

\paragraph{Cross-lingual Vocabulary Adaptation}
\citet{Tran2019FromET} used English BERT as a source LM. They initialized target language token representations as a weighted sum of the source embeddings followed by fine-tuning both the source and target models. \citet{wang-etal-2020-extending} and \citet{chau-etal-2020-parsing} added a fixed number of new target language tokens to the source vocabulary, expanding the source embedding matrix and output projection layers accordingly. The embeddings of the new tokens are randomly initialized over the expanded elements. Both studies performed additional pre-training on a target language corpus, often called language adaptive pre-training, i.e. LAPT \cite{chau-etal-2020-parsing}, after the target vocabulary initialization.
LAPT enables learning a target language model more efficiently than training it from scratch which is prohibitive with the size of current LLMs. It has become standard practice in more recent CVA studies~\cite{minixhofer-etal-2022-wechsel,Dobler2023FOCUSEE,downey-etal-2023-embedding,Ostendorff2023EfficientLM,Liu2023OFAAF}.
More recently, state-of-the-art methods completely replace the source embeddings with target language embeddings instead of expanding the source vocabulary~\cite{minixhofer-etal-2022-wechsel,Dobler2023FOCUSEE,Ostendorff2023EfficientLM,downey-etal-2023-embedding}. The aim is to utilize overlapping tokens between the source and target vocabularies for efficiency.

CVA has been extensively used to adapt generative LLMs to specific target languages~\cite{Cui2023EfficientAE,Balachandran2023TamilLlamaAN,Larcher2023CabritaCT,Pipatanakul2023TyphoonTL,Fujii2024ContinualPF}. However, the majority of these approaches simply expand the source embedding matrix followed by LAPT, while vocabulary replacement approaches have not been explored.
To the best of our knowledge, this is the first systematic study on the efficacy of various CVA methods for improving the inference efficiency of LLMs across languages.

\section{Cross-lingual Vocabulary Adaptation} \label{sec:method}

\subsection{Problem Setting} \label{subsec:problem}
Let $\mathcal{M}_\text{s}$ be a source pre-trained LLM with $\mathcal{T}_\text{s}$ and $\mathcal{V}_\text{s}$ its corresponding tokenizer and vocabulary. The aim is to learn a model $\mathcal{M}_\text{t}$ with the same architecture as $\mathcal{M}_\text{s}$ for a target language that supports a target vocabulary $\mathcal{V}_\text{t}$ given a tokenizer $\mathcal{T}_\text{t}$. $\mathcal{M}_\text{t}$ is first initialized with the weights of $\mathcal{M}_\text{s}$. Subsequently, its input embedding and output layer matrices are replaced such that the former is of dimensionality $|\mathcal{V}_\text{t}| \times H_\text{t}$ and the latter $H_\text{t} \times |\mathcal{V}_\text{t}|$, where $H_\text{t}$ is the hidden dimensionality of $\mathcal{M}_\text{t}$.
The weights of both matrices can be initialized by applying a target vocabulary initialization methods (\S\ref{subsec:method}).
Finally, $\mathcal{M}_\text{t}$ is adapted to the target language (i.e. with LAPT) by training it on target language data $\mathcal{D}$ using a causal language modeling objective.

\subsection{Target Vocabulary Initialization Methods} \label{subsec:method}
\paragraph{Random.}
The simplest approach is to randomly initialize the embeddings of $\mathcal{M}_\text{t}$~\cite{de-vries-nissim-2021-good,downey-etal-2023-embedding}.

\paragraph{Cross-lingual and Progressive Initialization (CLP).}
CLP~\citep{Ostendorff2023EfficientLM} first finds overlapping tokens between $\mathcal{V}_\text{t}$ and $\mathcal{V}_\text{s}$, i.e. $\mathcal{V}_\text{t} \cap \mathcal{V}_\text{s}$, and simply copies their weights from $\mathcal{M}_\text{s}$ to $\mathcal{M}_\text{t}$.
Each target token that does not overlap with any source token, i.e. $\mathcal{V}_\text{t} \backslash (\mathcal{V}_\text{t} \cap \mathcal{V}_\text{s})$ is initialized by its weighted average across all embeddings in $\mathcal{V}_\text{t} \cap \mathcal{V}_\text{s}$, i.e. common tokens in the source and target vocabularies. The weight of each embedding in $\mathcal{V}_\text{t} \cap \mathcal{V}_\text{s}$ is computed as the cosine similarity score between the respective overlapping token and the target non-overlapping token. Since there is no common representation between these two, CLP uses vector representations from an auxiliary target language-specific model ($\mathcal{M}_\text{aux}$) with the same tokenizer and vocabulary as $\mathcal{M}_\text{t}$.

\paragraph{Heuristics.}
\citet{downey-etal-2023-embedding} proposed a rule-based method for embedding initialization. %
First, embeddings are initialized according to their identity, in the same way that overlapping tokens are initialized in CLP, i.e. by copying from $\mathcal{M}_\text{s}$.
For all remaining tokens in $\mathcal{V}_\text{t}$, their embeddings are initialized based on the type of \textsc{script} identified by the Unicode block.
Each token that belongs to a particular script (e.g. Hebrew) is represented by a vector sampled from a Normal distribution with the same mean and standard deviation computed over all embeddings in $\mathcal{V}_\text{s}$ that belong to the same group.
A group can further be divided into two according to the \textsc{position} of each subword token, i.e. at the beginning or in the middle (e.g. ``\_the'' vs. ``the'').
Finally, the embeddings of any remaining tokens are randomly initialized.

\paragraph{FOCUS.}
\citet{Dobler2023FOCUSEE} proposed fast overlapping token combinations using sparsemax (FOCUS) initialization. Similar to CLP, FOCUS reuses the embeddings of $\mathcal{M}_\text{s}$ in $\mathcal{M}_\text{t}$ for tokens in $\mathcal{V}_\text{t} \cap \mathcal{V}_\text{s}$.
For non-overlapping tokens $\mathcal{V}_\text{t} \backslash (\mathcal{V}_\text{t} \cap \mathcal{V}_\text{s})$, it uses fastText~\cite{bojanowski-etal-2017-enriching} vectors trained on target specific data $\mathcal{D}$ tokenized by $\mathcal{T}_\text{t}$ 
to compute the cosine similarity between tokens in $\mathcal{V}_\text{t} \cap \mathcal{V}_\text{s}$ and $\mathcal{V}_\text{t} \backslash (\mathcal{V}_\text{t} \cap \mathcal{V}_\text{s})$.
It then applies sparsemax~\cite{pmlr-v48-martins16}, a sparse variant of softmax that assigns zero to any low-probability elements, over the similarity scores.
The token embeddings in $\mathcal{V}_\text{t} \backslash (\mathcal{V}_\text{t} \cap \mathcal{V}_\text{s})$ are finally initialized by taking the weighted sum of the source embeddings of tokens in $\mathcal{V}_\text{t} \cap \mathcal{V}_\text{s}$, where weights are the similarity scores with sparsemax applied.

\paragraph{CLP+.}
Finally, we propose CLP+, a modification to CLP motivated by the use of sparsemax in FOCUS. The aim is to dynamically select semantically similar tokens from $\mathcal{V}_\text{t} \cap \mathcal{V}_\text{s}$ to initialize a target embedding for a token in $\mathcal{V}_\text{t} \backslash (\mathcal{V}_\text{t} \cap \mathcal{V}_\text{s})$, leading to a better initialization of the embeddings \cite{Tran2019FromET}.
We follow the same process as CLP for tokens in $\mathcal{V}_\text{t} \cap \mathcal{V}_\text{s}$.
For non-overlapping tokens in $\mathcal{V}_\text{t} \backslash (\mathcal{V}_\text{t} \cap \mathcal{V}_\text{s})$, instead of taking the weighted average of \textit{all} overlapping source embeddings of $\mathcal{V}_\text{t} \cap \mathcal{V}_\text{s}$ as in CLP, we use the weighted sum of embeddings whose weight is calculated with sparsemax.
Note that the main difference between CLP+ and FOCUS is that the former uses $\mathcal{M}_\text{aux}$ while the latter uses fastText trained on $\mathcal{D}$ tokenized by $\mathcal{T}_\text{t}$ to compute similarities between $\mathcal{V}_\text{t} \cap \mathcal{V}_\text{s}$ and $\mathcal{V}_\text{t} \backslash (\mathcal{V}_\text{t} \cap \mathcal{V}_\text{s})$.\footnote{Note that proposing a new vocabulary initialization approach is not the main focus of our paper; we see CLP+ as a straightforward improvement of CLP motivated by FOCUS. Our overarching aim in this paper is to investigate how CVA methods affect inference efficiency and downstream performance of generative LLMs.}

\section{Experimental Setup} \label{sec:setup}

\subsection{Source Models} \label{subsec:model}
We use \textbf{BLOOM-1B} and \textbf{BLOOM-7B}~\cite{Scao2022BLOOMA1}, which are trained on data from 46 languages including Arabic (4.6\%) and Swahili (0.02\%).
We also use \textbf{TigerBot-7B}~\cite{chen2023tigerbot}, which is based on LLaMA 2~\cite{Touvron2023Llama2O} adapted using data from East Asian languages, i.e. Chinese (54\%), Korean (0.001\%), and Japanese (0.01\%).
Finally, we experiment with \textbf{Mistral-7B}~\cite{Jiang2023Mistral7} which is an English-centric model.
Table \ref{tab:tokenizers} shows the tokenizer and vocabulary size of each source model.
Note that the weights of the embedding and output layer matrices are tied for BLOOM but not for the rest.

\begin{table}[!t]
    \scriptsize
    \centering
    \begin{tabular}{l|cc}
        {Source ($\mathcal{M}_s$)} & Tokenizer ($\mathcal{T}_s$) & $|\mathcal{V}_s|$ \\
        \hline
        BLOOM & Byte-level BPE & 250,680\\
        TigerBot & Byte-level BPE & 60,512\\
        Mistral & Byte-level BPE & 32,000\vspace{3pt}\\
        
        {Target ($\mathcal{M}_t$)} & Tokenizer ($\mathcal{T}_t$) & $|\mathcal{V}_t|$  \\
        \hline
        German & Byte-level BPE & 50,257\\
        Japanese & Unigram & 32,000\\
        Arabic & Byte-level BPE & 64,000\\
        Swahili & Byte-level BPE & 50,257\\
    \end{tabular}
    \caption{Tokenizers and vocabulary size for source and target models.}
    \label{tab:tokenizers}
\end{table}

\subsection{Target Languages and Adaptation Data}
We experiment with a typologically diverse set of target languages including German (Indo-European), Japanese (Japonic), Arabic (Afro–Asiatic), and Swahili (Niger–Congo).
We use these languages because of the availability of language-specific (1) tokenizers; and (2) downstream task datasets with the same formulation across languages.\footnote{Note that data for the same task across languages does not match. Model performance is not directly comparable.} For adapting the source models, we use the OSCAR language-specific subcorpus~\cite{Jansen2022PerplexedBQ} for German, Arabic, and Japanese (Jan 2023).
For Swahili, we use its subset of CC-100~\cite{conneau-etal-2020-unsupervised} following \citet{minixhofer-etal-2022-wechsel}.
We use publicly available tokenizers and vocabularies for each target language. Table \ref{tab:tokenizers} shows the tokenizers and vocabulary size for the source and target models. More details are available in Table \ref{tab:tokenizer} in the Appendix.

\subsection{Tasks}
Following \citet{Ahia2023DoAL}, we experiment with four tasks in each target language with 500 test samples: (1) textual entailment (\textsc{nli}) using JNLI~\cite{kurihara-etal-2022-jglue} for Japanese and XNLI~\cite{conneau-etal-2018-xnli} for the rest; (2) X-CSQA~\cite{lin-etal-2021-common} for multiple choice question-answering (\textsc{mc}); (3) summarization (\textsc{sum}) including MLSUM~\cite{scialom-etal-2020-mlsum} for German and XL-Sum~\cite{hasan-etal-2021-xl} for the rest; and (4) span prediction (\textsc{span}) using XQuAD~\cite{artetxe-etal-2020-cross} for Arabic and German, JSQuAD~\cite{kurihara-etal-2022-jglue} for Japanese and KenSwQuAD~\cite{10.1145/3578553} for Swahili.\footnote{Due to computational constraints, we conduct zero-shot experiments on \textsc{sum}. For all other tasks, we evaluate models in zero- and few-shot settings. We use five demonstrations for \textsc{nli} and \textsc{mc} and three for \textsc{span} in the few-shot cases.}

\subsection{Prompt Templates}
We use the same English prompt templates as \citet{Ahia2023DoAL} for \textsc{nli} and \textsc{sum}.
For \textsc{mc} and \textsc{span}, we formulate a task-specific English prompt.
We translate the English prompt templates into each corresponding target language using Google Translate following \citet{yong-etal-2023-bloom}. The prompt templates can be found in Appendix \ref{appendix:prompt}.

\begin{figure*}[!t]
\begin{center}
\includegraphics[width=0.95\textwidth]{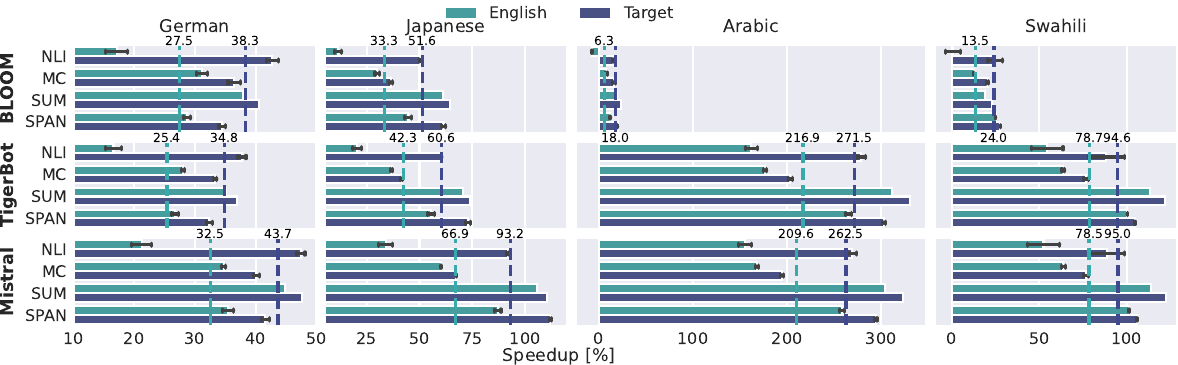}
\caption{
Relative speedup ratios to each base model/tokenizer when prompted in English and a target language.
Dotted lines denote the average speedup ratio across tasks in each setting.
}
\label{fig:speed}
\end{center}
\end{figure*}

\subsection{Baselines}
\label{subsec:baselines}
We compare the CVA methods against two baselines: (1) we use the source models directly on the target language tasks without any adaptation (\textbf{Source}); and (2) we adapt the source models by continuing pre-training on data from a target language by keeping the source vocabulary (\textbf{LAPT}) following \citet{yong-etal-2023-bloom}.

\subsection{Evaluation Metrics}

\paragraph{Inference Efficiency.} 
We calculate the average number of prompt tokens per sample for each dataset and tokenizer, and use its relative ratio to each source tokenizer as a proxy for inference speedup following \citet{Ahia2023DoAL} and \citet{petrov2023language}.
We use the average number of prompt tokens rather than the actual inference time because commercial APIs (e.g. OpenAI) often charge users on the basis of the total number of prompt and generated tokens. Note that inference efficiency is independent of the model size. Moreover, previous work \cite{petrov2023language,hong-etal-2024-accelerating} has shown a strong correlation between the length of tokenized inputs and actual processing times, i.e. longer input sequence leads to longer processing time.

\paragraph{Downstream Performance.}
For downstream performance evaluation, we use standard metrics for each dataset such as accuracy for \textsc{nli} and \textsc{mc}, F1 for \textsc{span}, and ROUGE-L~\cite{lin-2004-rouge} for \textsc{sum}.

\subsection{Implementation Details}
We perform our experiments under resource-constrained settings due to limited access to computational resources.
For computational efficiency, we use a low-rank adaptation approach LoRA~\cite{Hu2021LoRALA} applied on all linear layers (setting rank $r=8$) with LAPT, following \cite{yong-etal-2023-bloom,Cui2023EfficientAE,Balachandran2023TamilLlamaAN,Larcher2023CabritaCT}. 
We pre-train each model for a maximum of four days.
We use a batch size of 8 for BLOOM-1B and 16 for the 7B models with gradient accumulation steps set to 4 and a sequence length of 1,024.
We set the learning rate to 1e-4.
For a fair comparison, we use the checkpoints with the largest number of steps available across all vocabulary initialization approaches and the LAPT baseline for the same source model size and language.\footnote{For more details, see Appendix \ref{appendix:hyperparam}.}

\section{Results} \label{sec:results}

\subsection{Inference Efficiency} \label{sec:speedup}
Figure \ref{fig:speed} shows the relative inference speedup ratio between the target and source model prompted in the target language and English.
Overall, the results confirm our hypothesis that CVA accelerates inference in 95 out of 96 cases including zero- and few-shot settings.

Examining the efficiency of CVA across languages, we observe that the German CVA models show moderate average speedup ratios (25.4-43.7\%) across different tasks, source models and prompting languages. This is possibly due to the close relationship between German and English (i.e. both Germanic and Indo-European languages).
The Japanese target models also exhibit moderate but slightly greater average speedups compared to German of up to 60.6\% using BLOOM and TigerBot as source models.
In contrast, inference speedups are substantially greater using Mistral as the source model (66.9-93.2\% on average).
These differences may stem from the inclusion of Chinese pre-training data in BLOOM,\footnote{Note that the Japanese script includes Chinese characters.} and Chinese and Japanese data in TigerBot.
Arabic and Swahili target models obtain smaller speedups than the other languages using BLOOM as the source model (up to 24.0\% on average). This is also likely due to the inclusion of Arabic and Swahili pre-training data in BLOOM.
In contrast, CVA models in both languages obtain substantial gains using TigerBot and Mistral, up to an impressive 271.5\% for Arabic and 95.0\% for Swahili.
This is due to the absence of the two languages from the training data of TigerBot and Mistral\footnote{We do not have enough information about the training data of Mistral apart from that it is \href{https://mistral.ai/news/announcing-mistral-7b/}{good in English tasks}.}, and the different Arabic script.

\setlength{\tabcolsep}{5.4pt}
\renewcommand*{\arraystretch}{1.0}
\begin{table*}[!t]
\begin{center}
\tiny
\resizebox{\linewidth}{!}{%
\begin{tabular}{c|lllll|llll|llll|llll}
&& \multicolumn{4}{c|}{\textbf{German}}  & \multicolumn{4}{c|}{\textbf{Japanese}} & \multicolumn{4}{c|}{\textbf{Arabic}} & \multicolumn{4}{c}{\textbf{Swahili}}\\

&\textbf{Model} & \multicolumn{1}{c}{\scriptsize \textsc{nli}} & \multicolumn{1}{c}{\scriptsize \textsc{mc}} & \multicolumn{1}{c}{\scriptsize \textsc{sum}} & \multicolumn{1}{c|}{\scriptsize \textsc{span}} 
 & \multicolumn{1}{c}{\scriptsize \textsc{nli}} & \multicolumn{1}{c}{\scriptsize \textsc{mc}} & \multicolumn{1}{c}{\scriptsize \textsc{sum}} & \multicolumn{1}{c|}{\scriptsize \textsc{span}} 
 & \multicolumn{1}{c}{\scriptsize \textsc{nli}} & \multicolumn{1}{c}{\scriptsize \textsc{mc}} & \multicolumn{1}{c}{\scriptsize \textsc{sum}} & \multicolumn{1}{c|}{\scriptsize \textsc{span}} 
& \multicolumn{1}{c}{\scriptsize \textsc{nli}} & \multicolumn{1}{c}{\scriptsize \textsc{mc}} & \multicolumn{1}{c}{\scriptsize \textsc{sum}} & \multicolumn{1}{c}{\scriptsize \textsc{span}} \\

\hline
\multirow{19}{*}{\rotatebox[origin=c]{90}{\parbox[c]{1cm}{\centering \textbf{Zero-shot}}}}
& \cellcolor{gray!25}\textbf{BLOOM-1B} (Source) & \cellcolor{gray!25}.35 & \cellcolor{gray!25}.21 & \cellcolor{gray!25}17.8 & \cellcolor{gray!25}.06 & \cellcolor{gray!25}.29 & \cellcolor{gray!25}.20 & \cellcolor{gray!25}18.2 & \cellcolor{gray!25}.22 & \cellcolor{gray!25}.31 & \cellcolor{gray!25}.20 & \cellcolor{gray!25}12.0 & \cellcolor{gray!25}.15 & \cellcolor{gray!25}.35 & \cellcolor{gray!25}.22 & \cellcolor{gray!25}12.0 & \cellcolor{gray!25}.03\\
&\cellcolor{gray!25}\textbf{BLOOM-1B} (LAPT) & \cellcolor{gray!25}.34 & \cellcolor{gray!25}.22 & \cellcolor{gray!25}14.3 & \cellcolor{gray!25}.09 & \cellcolor{gray!25}.28 & \cellcolor{gray!25}.20 & \cellcolor{gray!25}20.7 & \cellcolor{gray!25}.26 & \cellcolor{gray!25}.31 & \cellcolor{gray!25}.19 & \cellcolor{gray!25}11.4 & \cellcolor{gray!25}.13 & \cellcolor{gray!25}.35 & \cellcolor{gray!25}.18 & \cellcolor{gray!25}7.7 & \cellcolor{gray!25}.07\\
\hline
\end{tabular}%
}
\caption{Mean performance over five runs with in-language prompting on 500 randomly selected test samples for each dataset.
\colorbox{gray!25}{Gray} denotes baselines without CVA.
\textbf{Bold} indicates comparable or better results than the baselines.
Darker \textcolor{blue}{blue} and \textcolor{red}{red} indicate higher positive and negative relative performance change over Source, respectively.
}
\label{tab:main_result}
\end{center}
\end{table*}

Looking into individual tasks,
we observe that CVA models gain larger speedup ratios in \textsc{span} and \textsc{sum} compared to the other two tasks across LLMs and languages.
In particular, we record a maximum speedup of 331\% in Arabic \textsc{sum} with in-language prompting using TigerBot as source.
In contrast, speedup ratios tend to be smaller than average in \textsc{nli} and \textsc{mc} across different target and source models, except for \textsc{nli} with in-language prompting.
Specifically, the Arabic model using BLOOM as source shows a slowdown of 7.63\% when prompted in English.
Our hypothesis is that this is due to the ratio of English-related words included in a prompt in each task, resulting to overfragmentation of such words by the \textit{target-language} tokenizer, which is detrimental to inference speedup.
Indeed, the number of tokens of the \textsc{nli} English prompt template\footnote{\texttt{Question: True, False, or Neither? Answer:}} is ten when tokenized with the BLOOM source tokenizer, increasing to 21 with the Arabic tokenizer.

Finally, we investigate the inference efficiency for the target models by prompting language.
We observe that the target models show greater inference speedup ratios with in-language prompts than English in all cases.
The average differences between in-language and English prompts are 12.8\%, 24.6\%, and 26.8\%, using BLOOM, TigerBot, and Mistral as source models, respectively.
This suggests that the CVA models are susceptible to code-mixed text (i.e. including English prompts), leading to overfragmentation for words not in a target language.
Furthermore, in-language prompting is a more realistic scenario for non-English speakers to use LLMs than English prompting.
These differences highlight the advantage of CVA and confirm the limitations of using a source tokenizer, reported by \citet{Ahia2023DoAL} and \citet{petrov2023language}.

\begin{figure*}[t]
\begin{center}
\includegraphics[width=\textwidth]{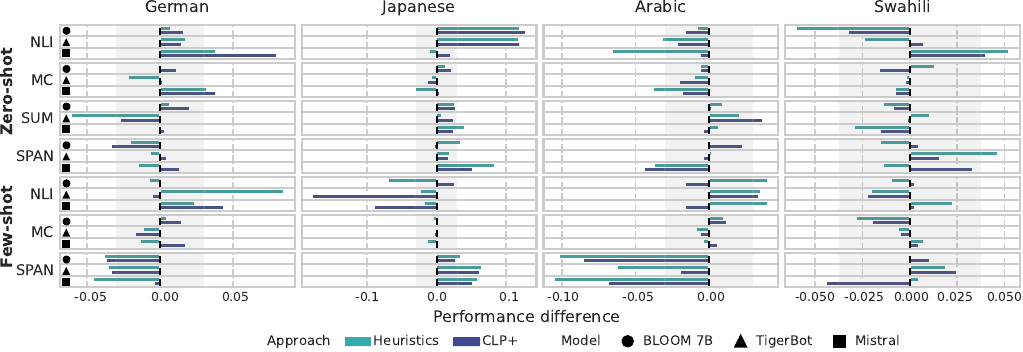}
\caption{
Performance difference between English and in-language prompts. Positive and negative values indicate better performance using in-language or English prompts respectively. 
}
\label{fig:prompt-preference}
\end{center}
\end{figure*}

\subsection{Downstream Performance} \label{subsec:performance}
We compare the downstream performance of all CVA methods (\S\ref{subsec:method}) to the Source and LAPT baselines (\S\ref{subsec:baselines}).
Table \ref{tab:main_result} shows the zero- and few-shot performance of all models with in-language prompting.\footnote{Full results are in Table \ref{tab:main_result_target} in the Appendix. We only show results with Heuristics and CLP+ for 7B models in Table \ref{tab:main_result} as two representative approaches for brevity.}
Results using English prompts are included in Table \ref{tab:main_result_full} in the Appendix. %

Overall, CVA models show comparable or better performance than the baselines in the majority of the cases across tasks and languages using BLOOM-1B as source.
Models adapted with simple Random target vocabulary initialization are competitive compared to more sophisticated approaches and the baselines in the majority of the cases -- 17 for Source and 26 for LAPT out of 28 cases, respectively.
However, they are not as robust with English prompting (see Appendix \ref{subsec:additional_result}).
CVA with Heuristics also performs similar to the semantic similarity-based methods (i.e. CLP, FOCUS and CLP+) corroborating findings by \citet{downey-etal-2023-embedding}.
They are similar to or better than Source in 18 out of 28 cases, and LAPT in 20 out of 28 cases without a substantial drop in performance observed in Random with English prompting.

Experiments with larger source models show that adapting BLOOM-7B with Heuristics is generally on par with CLP+ and other semantic similarity-based methods. We also note that it outperforms Source and LAPT in 15 and 19 out of 28 cases across tasks and languages, respectively.
When using TigerBot-7B and Mistral-7B as source, we find that CVA models are not always better than the baselines.
For instance, CLP+ exhibits similar performance to Source and LAPT in 15 and 14 out of 28 cases, while Heuristics is similar to or better than them in 13 and 14 cases, respectively.
This suggests that LLMs such as TigerBot and Mistral, which are not as multilingual as BLOOM, may not perform well with CVA. This is possibly due to less transferable cross-lingual knowledge, and the small amount of target language data included during pre-training.
Furthermore, Heuristics CVA might not be suitable with such LLMs, especially for languages that are not included in pre-training (i.e. Arabic and Swahili for TigerBot and Mistral), as we observe poor performance in generative tasks like \textsc{sum} and \textsc{span} in Arabic and Swahili compared to CLP+.

Finally, we observe that language overlaps between source and target models affect downstream performance, similar to inference speedups, especially in generative tasks such as \textsc{sum} and \textsc{span}.
BLOOM-based CVA models show substantial performance improvement in German and Japanese zero-shot \textsc{span} compared to baselines, while they generally achieve competitive or slightly lower performance in Arabic and Swahili across tasks, except for Swahili zero-shot \textsc{span} in BLOOM-1B.
CVA models based on TigerBot-7B and Mistral-7B  with CLP+ for Arabic and Swahili are competitive to the baselines across generative tasks (3 out of 4 cases in zero-shot tasks regardless of the source model). However, this is not the case for German and Japanese CVA models.\footnote{TigerBot is built on top of LLaMA2, with German being the largest among non-English languages in the pre-training corpus (0.17\%)~\cite{Touvron2023Llama2O}.}

\section{Analysis} \label{sec:analysis}

\paragraph{In-language vs. English Prompting.} \label{subsec:prompt_analysis}

Figure \ref{fig:prompt-preference} shows the performance difference between English and in-language prompting across models, languages and tasks for two representative CVA methods (Heuristics and CLP+).\footnote{One heuristic- and one semantic similarity-based method.}

Recent studies have showed that in-language prompting yields lower performance than prompting in English~\cite{lin-etal-2022-shot,ahuja-etal-2023-mega,muennighoff-etal-2023-crosslingual}.
Surprisingly, we find no major performance drop with in-language prompting in the majority of the zero-shot settings across languages.
We note similar or better performance with in-language prompting in 11 out of 16 cases. 
We also observe a drop of 0.03 or larger (non-shaded areas in Figure \ref{fig:prompt-preference}) in only 6 out of 16 cases.
The few-shot settings also exhibited similar trends with substantial performance degradation of 0.03 or more in 4 out of 12 cases, and similar or better performance in 7 out of 12 cases.
Some in-language prompting cases with lower performance than English, such as in German across tasks and zero-shot \textsc{nli} in Arabic and Swahili, can be related to the tokenization effects discussed in \S\ref{sec:speedup}. Previous studies have also found a strong correlation between tokenization and performance~\cite{rust-etal-2021-good,bostrom-durrett-2020-byte,fujii-etal-2023-different}.

\begin{figure}[t!]
\begin{center}
\includegraphics[width=\columnwidth]{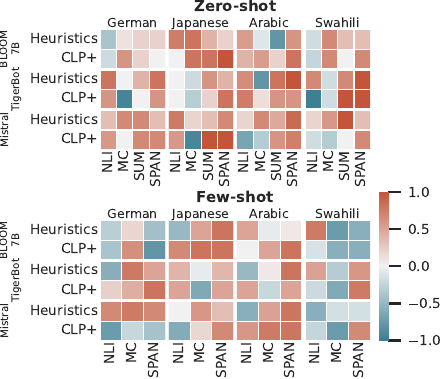}
\caption{
Kendall's $\tau$ correlation between the number of LAPT steps and performance (in-language prompting).
}
\label{fig:step_ablation}
\end{center}
\end{figure}

\paragraph{LAPT Steps.}
LAPT is an integral part of CVA~\cite{minixhofer-etal-2022-wechsel,Dobler2023FOCUSEE,Ostendorff2023EfficientLM,downey-etal-2023-embedding}. However, it is computationally intensive, requiring updating models over a large number of training steps. Therefore, we investigate the relationship between downstream performance and the number of LAPT steps.\footnote{Every 2k steps starting from 1k and every 10k after 13k.}
Figure \ref{fig:step_ablation} shows the Kendall's tau~\cite{665905b2-6123-3642-832e-05dbc1f48979} correlation coefficients between them.
Overall, LAPT helps improve downstream performance in both zero- and few-shot settings in 69.5\% and 59.4\% of the cases, respectively.\footnote{We observe similar trends with English prompting (Figure \ref{fig:step_ablation_target} in the Appendix).}
In particular, both TigerBot and Mistral, which are not as multilingual as BLOOM, tend to benefit more from LAPT, especially in zero-shot \textsc{sum} and \textsc{span} across languages. This suggests that LAPT helps CVA models to increase target language knowledge and reach a performance similar BLOOM-7B (Source) in a similar number of steps.

Examining the correlation by task across zero- and few-shot settings, we often observe negative or no correlation in classification tasks like \textsc{nli} and \textsc{mc}.
In contrast, \textsc{span} and \textsc{sum} generally benefit from LAPT in zero-shot settings across languages, in addition to few-shot \textsc{span} in Japanese and Arabic.
We hypothesize that zero-shot generative tasks, i.e. \textsc{sum} and \textsc{span}, can be more challenging than the other text classification tasks~\cite{davletov-etal-2021-liori,yamaguchi-etal-2022-hitachi}, and thus require better target language representations to perform well.

\begin{figure}[t!]
\begin{center}
\includegraphics[width=\columnwidth]{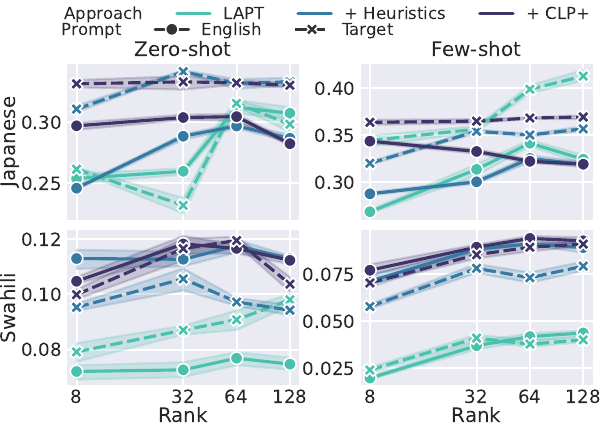}
\caption{Performance changes in \textsc{span} with respect to LoRA rank $r$.
}
\label{fig:rank_ablation}
\end{center}
\end{figure}

\paragraph{LoRA Rank $r$.} \label{subsec:rank}
There is a trade-off between computational efficiency and performance when adapting LLMs with LoRA~\cite{Hu2021LoRALA}. We further analyze how the LoRA rank affects performance in CVA. 
To keep computational costs low, we experiment by setting $r = \{8, 32, 64, 128\}$ using BLOOM-1B on \textsc{span} in Japanese and Swahili where we observe large performance variations (Table \ref{tab:main_result}).
Figure \ref{fig:rank_ablation} shows how performance changes with respect to $r$.
On the one hand, the performance of CVA models does not generally increase with $r$ in the zero-shot setting.
On the other hand, performance improves with $r$ in the few-shot setting. %
This suggests that setting $r=8$ is a reasonable choice in zero-shot settings.
Increasing $r$ to 32, 64 or 128 can yield better few-shot performance but results to higher computational costs.
For instance, the best-performing Swahili model ($r=64$) results in a 14\% increase in the number of trainable parameters compared to $r=8$.

\section{Recommendations} \label{sec:recommendation}
Our findings suggest that CVA offers substantial inference speedups across tasks, languages and models regardless of the target embedding initialization approach. However, downstream performance is sensitive to the embedding initialization and LoRA fine-tuning. Therefore, we provide the following recommendations to researchers and practitioners\footnote{If we do not care about inference efficiency at all, it might make sense to use LAPT instead of CVA methods.}:
\begin{enumerate}[left=0pt]
    \item  Simple Heuristics-based initialization should be used to save computational costs when the source model is sufficiently multilingual, covering several languages and scripts. This should also be the case when the target language is included in the source model pre-training data.
    \item Semantic similarity-based initialization methods (e.g. FOCUS, CLP, CLP+) should be used in cases where the target language is not included in the source model pre-training data, to obtain better downstream performance.
    \item Careful cost-benefit consideration is needed to choose an optimal LoRA rank $r$. A low rank ($r=8$) is a good starting point in zero-shot settings considering performance and computational costs; $ r=32$ or larger is recommended in few-shot settings.    
\end{enumerate}

\section{Conclusion} \label{sec:conclusion}
We have conducted an extensive study on the effectiveness of CVA on LLM inference efficiency and performance.
Our experiments in four diverse languages demonstrated that CVA substantially contributes to LLM inference speedups of up to 271.5\% while maintaining comparable downstream performance to baselines when adapting multilingual LLMs. We supplement our results and analysis with specific recommendations for effective CVA with LLMs.
In future work, we plan to explore various inference-aware methods for cross-lingual transfer, such as cost-effective subword vocabulary selection~\cite{gee-etal-2023-multi}.

\section*{Limitations}
\paragraph{Prompt Tuning.} 
We use a translated version of in-language prompts from English. This may affect the downstream performance due to machine translation noise, underestimating the performance of in-language prompting.

\paragraph{Languages.}
This study covers four linguistically diverse languages (German, Arabic, Japanese, and Swahili), following previous work on CVA that has also tested a similar number of languages.
For instance, \citet{de-vries-nissim-2021-good} tested two languages, and \citet{Dobler2023FOCUSEE} tested five languages.
Nonetheless, exploring more languages is an interesting avenue for future work but out of the scope of this paper given our limited computing capacity.

\paragraph{Model Size.}
We use LLMs of various sizes ranging from 1B to 7B, which are far larger than those tested in previous CVA studies. 
For example, \citet{Dobler2023FOCUSEE}, \citet{Liu2023OFAAF}, and \citet{downey-etal-2023-embedding} use XLM-R~\cite{conneau-etal-2020-unsupervised} (0.28B).
Note that inference efficiency measured by the number of processed (or generated) tokens is not affected by the model size. However, investigating the performance of CVA approaches with larger models would be valuable in future studies.

\section*{Acknowledgments}
We would like to thank Ahmed Alajrami for proofreading Arabic prompts, and Olga Iakovenko, Constantinos Karouzos, and Huiyin Xue for their valuable feedback.
AY is supported by the Engineering and Physical Sciences Research Council (EPSRC)  [grant number EP/W524360/1] and the Japan Student Services Organization (JASSO) Student Exchange Support Program (Graduate Scholarship for Degree Seeking Students).
This work is also partly supported by the EPSRC project EP/T02450X/1.

\bibliography{anthology,custom}

\clearpage
\appendix

\section*{Appendix}
\section{Implementation Details}
\subsection{Tokenizer} 
\label{appendix:tokenizer}
To reduce the computational costs, we utilized publicly available existing tokenizers for each target language, which means we used them as $\mathcal{T}_\text{t}$.
Table \ref{tab:tokenizer} lists the tokenizers used in our experiments.

\begin{table*}[t]
\small
\begin{center}
\begin{tabular}{lllll}
\textbf{Language} & \begin{tabular}[c]{@{}c@{}}\textbf{Tokenization}\\ \textbf{Algorithm}\end{tabular} &  \textbf{Hugging Face Identifier} & \textbf{Citation} & \textbf{License}\\ 
\hline
German & Byte-level BPE  & \href{https://huggingface.co/malteos/gpt2-xl-wechsel-german}{malteos/gpt2-xl-wechsel-german} & & MIT\\
Japanese & Unigram  & \href{https://huggingface.co/rinna/japanese-gpt-neox-3.6b-instruction-ppo}{rinna/japanese-gpt-neox-3.6b-instruction-ppo} & & MIT\\
Arabic & Byte-level BPE  & \href{https://huggingface.co/aubmindlab/aragpt2-base}{aubmindlab/aragpt2-base} & \cite{antoun-etal-2021-aragpt2} & \href{https://raw.githubusercontent.com/aub-mind/arabert/master/aragpt2/LICENSE}{See here} \\
Swahili & Byte-level BPE  & \href{https://huggingface.co/benjamin/gpt2-wechsel-swahili}{benjamin/gpt2-wechsel-swahili} & \cite{minixhofer-etal-2022-wechsel} & MIT\\ \hline
\end{tabular}
\caption{List of tokenizers used for each language-specific model with vocabulary adaptation.}
\label{tab:tokenizer}
\end{center}
\end{table*}

\subsection{Language-specific pre-trained LM}
For language-specific pre-trained LMs used in CLP and CLP+, we used the corresponding models to $\mathcal{T}_\text{t}$, which are listed in Table \ref{tab:tokenizer} and are all decoder-based models.
Note that we only used the embedding of each language-specific pre-trained LM for vocabulary adaptation, and therefore, one can also use encoder and encoder-decoder based models.

\subsection{fastText in FOCUS}
For FOCUS, we trained a fastText model for each language on a corresponding CC-100~\cite{conneau-etal-2020-unsupervised} text with the same configuration as \citet{Dobler2023FOCUSEE}.

\subsection{Hyperparameters and Generation Configurations} \label{appendix:hyperparam}
\paragraph{LAPT}
Table \ref{tab:hyperparams_pretraining} shows the hyperparameters in LAPT for each model size.
Note that due to the computational resource constraints and funds for running experiments, we could run pre-training of up to four days for each approach.
Therefore, we picked up checkpoints with the largest number of steps available across models with the same base model (i.e. BLOOM-1B, BLOOM-7B, etc.) and language for evaluation to make a fair comparison.
We also temporarily trimmed the unused embeddings of BLOOM models for LAPT, whose tokens did not appear in the training corpus during pre-training to save memory and for faster computation.\footnote{We used the implementations by \citet{Ushio2023AnEM} and \citet{Williams2023FrustratinglySM}.}

\begin{table}[h]
\begin{center}
\small
\begin{tabular}{lcc}
\textbf{Hyperparameters} & \textbf{1B} & \textbf{7B} \\
\hline
Batch size & 8 & 16\\
Gradient accumulation steps & 4 & 4\\
Maximum number of training epochs & 1 & 1\\
Maximum number of training days & 4 & 4\\
Adam $\epsilon$ & 1e-8 & 1e-8\\
Adam $\beta_1$ & 0.9 & 0.9\\
Adam $\beta_2$ & 0.999 & 0.999\\
Sequence length & 1,024 & 1,024\\
Learning rate & 1e-4 & 1e-4\\
Learning rate scheduler & cosine & cosine\\
Warmup steps & 100 & 100 \\
Weight decay & 0.01 & 0.01\\
Attention dropout & 0.0 & 0.0 \\
Dropout & 0.05 & 0.05\\
LoRA rank $r$ & 8 & 8\\
LoRA dropout & 0.05 & 0.05\\
LoRA $\alpha$ & 32 & 32 \\
Training precision & FP16 & FP16\\
Model quantization & int 8 & int 8\\
\hline
\end{tabular}%
\caption{Hyperparameters for LAPT.}
\label{tab:hyperparams_pretraining}
\end{center}
\end{table}

\paragraph{Generation}
Following \citet{Cui2023EfficientAE}, we introduced a verbalizer for the classification tasks: \textsc{nli} and \textsc{mc}, where we mapped the first generated token into a label to compute accuracy.
For mapping, we simply picked up a token with the maximum log-likelihood among candidate tokenized words.
The list of candidate label words for each task is shown in Table \ref{tab:label}.
Table \ref{tab:params_generation} lists the parameters used during evaluation.
To make a fair comparison, we did not conduct any generation parameter tuning and used the same ones across all approaches.
For \textsc{sum} and few-shot \textsc{span} in Swahili, we truncated an article whenever it exceeded the maximum prompt length of 4,096 to avoid the CUDA out-of-memory error.

\begin{table}
    \centering
    \small
    \begin{tabular}{ccc}
       \textbf{Task}  & \textbf{Language}  & \textbf{Label words}\\
    \hline
        \textsc{nli} & English & True, False, Neither\\
         & German & Wahr, Falsch, Weder\\
         & Japanese & \begin{CJK}{UTF8}{min}真\end{CJK}, \begin{CJK}{UTF8}{min}偽\end{CJK}, \begin{CJK}{UTF8}{min}どちらでもない\end{CJK}\\
         & Arabic & \RL{صحيح}, \RL{خطأ}, \RL{لا هذا ولا ذاك}\\
         & Swahili & Kweli, Uongo, Wala\\
    \hline
        \textsc{mc} & All & A, B, C, D, E\\
    \hline
    \end{tabular}
    \caption{List of candidate label words for each classification task.}
    \label{tab:label}
\end{table}

\begin{table}[h]
\begin{center}
\begin{tabular}{lc}
\textbf{Parameters} & \textbf{Values} \\
\hline
Maximum prompt length & 4,096\\
Temperature & 0.8\\
Repetition penalty & 1.1\\
Top $k$ & 40\\
Top $p$ & 0.9\\
Beam width & 5\\
Sampling & True\\
Early stopping & True\\
\hline
\end{tabular}%
\caption{Parameters for generation.}
\label{tab:params_generation}
\end{center}
\end{table}

\subsection{Checkpoints}
As explained in \ref{appendix:hyperparam}, we trained all models for up to four days each due to limited computational resources and funds for experiments.
The only exception was Swahili since the dataset is small enough to complete LAPT.
To make a fair comparison, we used checkpoints with the largest number of steps available across models with the same target language and base model.
Table \ref{tab:checkpoint} shows the list of checkpoints used for evaluation.

\begin{table}[t!]
\begin{center}
\small
\begin{tabular}{lllll}
\textbf{Model} & \multicolumn{4}{c}{\textbf{Language}}\\
 & \multicolumn{1}{c}{\scriptsize de} & \multicolumn{1}{c}{\scriptsize ja} & \multicolumn{1}{c}{\scriptsize ar} & \multicolumn{1}{c}{\scriptsize sw}\\
\hline
BLOOM-1B & 47k & 48k & 50k & 9k\\
BLOOM-7B & 8k & 8k & 8k & 4k\\
TigerBot-7B & 8k & 8k & 8k & 4k\\
Mistral-7B & 6k & 6k & 6k & 4k\\
\hline
\end{tabular}%
\caption{List of checkpoints used for evaluation.
We used checkpoints with the largest number of steps available across all models with the same base model and language.
}
\label{tab:checkpoint}
\end{center}
\end{table}

\subsection{Libraries and Hardware}
We implement our models using PyTorch~\cite{NEURIPS2019_9015}, Hugging Face Transformers~\cite{wolf-etal-2020-transformers} and PEFT~\cite{peft}.
We preprocess data with Hugging Face Datasets~\cite{lhoest-etal-2021-datasets}.
For evaluation, we use Hugging Face Evaluate\footnote{\url{https://github.com/huggingface/evaluate}} to compute downstream performance metrics.
We use a single NVIDIA A100 (80GB) GPU for all experiments.

\subsection{Prompt Templates}
\label{appendix:prompt}
Table \ref{tab:prompt} shows the prompt templates used in our evaluation.

\subsection{Code}
Our code and models are available here: \url{https://github.com/gucci-j/llm-cva}.

\begin{table*}[t]
\begin{center}
\begin{tabular}{lll}
\textbf{Task} & \textbf{Language} & \textbf{Template} \\ 
\hline
\multirow{5}{*}{\textsc{nli}} & English & \{premise\} Question: \{hypothesis\} True, False, or Neither? Answer: \\
 & German & \{premise\} Frage: \{hypothesis\} Wahr, Falsch oder Weder? Antwort: \\
 & Japanese & \{premise\} \begin{CJK}{UTF8}{min}質問\end{CJK}: \{hypothesis\} \begin{CJK}{UTF8}{min}真、偽、どちらでもない？\end{CJK} \begin{CJK}{UTF8}{min}答え\end{CJK}: \\
 & Arabic & \{premise\} \RL{سؤال}: \{hypothesis\} \RL{صحيح ، خطأ أو لا هذا ولا ذاك؟ إجابة}: \\
 & Swahili & \{premise\} Swali: \{hypothesis\} Kweli, Uongo au Wala? Jibu: \\ 

\hline
\multirow{5}{*}{\textsc{mc}} & English & \begin{tabular}[c]{@{}l@{}}\{question\} A. \{choice\_1\}, B. \{choice\_2\}, C. \{choice\_3\}, D. \{choice\_4\}, \\ E. \{choice\_5\} Answer:\end{tabular} \\
 & German & \begin{tabular}[c]{@{}l@{}}\{question\} A. \{choice\_1\}, B. \{choice\_2\}, C. \{choice\_3\}, D. \{choice\_4\}, \\E. \{choice\_5\} Antwort:\end{tabular} \\
 & Japanese & \begin{tabular}[c]{@{}l@{}}\{question\} A. \{choice\_1\}, B. \{choice\_2\}, C. \{choice\_3\}, D. \{choice\_4\}, \\E. \{choice\_5\} \begin{CJK}{UTF8}{min}答え\end{CJK}:\end{tabular} \\
 & Arabic & \begin{tabular}[c]{@{}l@{}}\{question\} A. \{choice\_1\}, B. \{choice\_2\}, C. \{choice\_3\}, D. \{choice\_4\}, \\E. \{choice\_5\} \RL{إجابة}:\end{tabular} \\
 & Swahili & \begin{tabular}[c]{@{}l@{}}\{question\} A. \{choice\_1\}, B. \{choice\_2\}, C. \{choice\_3\}, D. \{choice\_4\}, \\E. \{choice\_5\} Jibu:\end{tabular} \\

\hline
\multirow{5}{*}{\textsc{sum}} & English & \begin{tabular}[c]{@{}l@{}}Write a short summary of the following text in \{language\}. \\Article: \{text\} Summary:\end{tabular} \\
 & German & \begin{tabular}[c]{@{}l@{}}Schreiben Sie eine kurze Zusammenfassung des folgenden Textes auf Deutsch. \\Artikel: \{text\} Zusammenfassung:\end{tabular} \\
 & Japanese & \begin{CJK}{UTF8}{min}次の文章の要約を日本語で書きなさい。記事\end{CJK}: \{text\} \begin{CJK}{UTF8}{min}要約\end{CJK}: \\
 & Arabic & \RL{اكتب ملخصًا قصيرًا للنص التالي باللغة العربية. المقالة}: \{text\} \RL{الملخص}: \\
 & Swahili & \begin{tabular}[c]{@{}l@{}}Andika muhtasari mfupi wa maandishi yafuatayo kwa Kiswahili. \\Makala: \{text\} Muhtasari:\end{tabular} \\

\hline
\multirow{5}{*}{\textsc{span}} & English & \begin{tabular}[c]{@{}l@{}}Answer the following question. Context: \{context\} Question: \{question\} \\Answer:\end{tabular}  \\
 & German & \begin{tabular}[c]{@{}l@{}}Beantworten Sie die folgende Frage. Artikel: \{context\} Frage: \{question\} \\ Antwort:\end{tabular} \\
 & Japanese & \begin{tabular}[c]{@{}l@{}}\begin{CJK}{UTF8}{min}次の文章の質問に答えなさい。文章\end{CJK}: \{context\} \begin{CJK}{UTF8}{min}質問\end{CJK}: \{question\}\\ \begin{CJK}{UTF8}{min}答え\end{CJK}:\end{tabular} \\
 & Arabic & \RL{أجب على السؤال التالي. سياق}: \{context\} \RL{السؤال}: \{question\} \RL{الإجابة}: \\
 & Swahili & Jibu swali lifuatalo. Makala: \{context\} Swali: \{question\} Jibu: \\
 
 \hline
\end{tabular}
\caption{Prompt template for each task and language.}
\label{tab:prompt}
\end{center}
\end{table*}

\clearpage
\section{Licenses}
This study used various publicly available models and datasets with different licenses, as detailed below, all of which permit their use for academic research.

\subsection{Models}
BLOOM is licensed under the BigScience RAIL License.\footnote{\url{https://huggingface.co/spaces/bigscience/license}}
TigerBot and Mistral are licensed under the Apache-2.0 License.
The licenses of the helper models are listed in Table \ref{tab:tokenizer}.

\subsection{Datasets}
XNLI is distributed under CC BY-NC 4.0.
JNLI, XQuAD, and JSQuAD are distributed under CC BY-SA 4.0.
XCSQA is a derivative of CommonsenseQA \cite{talmor-etal-2019-commonsenseqa}, which is licensed under an MIT license.
OSCAR and KenSwQuAD are licensed under CC0 -- no rights reserved.
XL-Sum is licensed under CC BY-NC-SA 4.0, while MLSUM is distributed under an MIT license.

\section{Results} \label{sec:appendix_result}
\subsection{Perplexity}
Table \ref{tab:perplexity} shows the perplexities of adapted models measured on the language-specific subset of Wikipedia 50K articles.

\begin{table}[t]
    \small
    \centering
    \begin{tabular}{lllll}

    \textbf{Model} & \textbf{German} & \textbf{Japanese} & \textbf{Arabic} & \textbf{Swahili}\\

\hline
\multicolumn{5}{l}{\textbf{BLOOM-1B}}\\
\rowcolor{gray!25}
Source & 45.6 & 44.7 & 14.6 & 45.4\\
\rowcolor{gray!25}
LAPT & 22.6 & 21.0 & 20.9 & 55.4\\
Random & \textbf{58.5} & 55.1 & 53.7 & 305.1\\
CLP & 75.2 & 62.6 & 46.5 & 190.2\\
Heuristics & 71.5 & 51.2 & \underline{46.0} & \underline{176.5}\\
FOCUS & \underline{70.8} & \underline{50.1} & 46.1 & \textbf{168.9}\\
CLP+ & 75.5 & \textbf{48.8} & \textbf{45.6} & 185.5\\

\hline
\multicolumn{5}{l}{\textbf{BLOOM-7B}}\\
\rowcolor{gray!25}
Source & 18.7 & 21.4 & 9.5 & 14.9\\
\rowcolor{gray!25}
LAPT & 13.6 & 13.7 & 11.0 & 19.4\\
Random & 45.0 & 54.1 & 44.5 & 179.1\\
CLP & 49.1 & 165.9 & \underline{30.9} & 91.8\\
Heuristics & \textbf{36.4} & 43.5 & 32.0 & \underline{85.8}\\
FOCUS & \underline{36.8} & \underline{42.3} & 32.0 & \textbf{82.4}\\
CLP+ & 37.0 & \textbf{41.8} & \textbf{29.9} & 89.7\\

\hline
\multicolumn{5}{l}{\textbf{TigerBot-7B}}\\
\rowcolor{gray!25}
Source & 7.6 & 8.2 & 5.1 & 30.9\\
\rowcolor{gray!25}
LAPT & 8.6 & 10.0 & 3.0 & 9.5\\
Random & 116.2 & 77.8 & 130.7 & 636.3\\
CLP & 87.6 & 39.9 & 103.0 & 688.3\\
Heuristics & 47.3 & 41.4 & 124.1 & 605.4\\
FOCUS & \textbf{43.9} & \underline{39.2} & \underline{81.8} & \textbf{398.6}\\
CLP+ & \underline{45.2} & \textbf{38.8} & \textbf{81.6} & \underline{559.4}\\

\hline
\multicolumn{5}{l}{\textbf{Mistral-7B}}\\
\rowcolor{gray!25}
Source & 4.5 & 8.0 & 3.7 & 18.4\\
\rowcolor{gray!25}
LAPT & 5.5 & 9.7 & 2.9 & 7.5\\
Random & 78.0 & 54.4 & 77.2 & 587.7\\
CLP & \underline{33.7} & 41.4 & 74.5 & 358.6\\
Heuristics & 34.8 & 41.4 & 81.5 & 413.3\\
FOCUS & \textbf{33.4} & \underline{41.0} & \underline{66.1} & \textbf{287.4}\\
CLP+ & 35.4 & \textbf{40.0} & \textbf{58.1} & \underline{297.0}\\
\hline
    \end{tabular}
    \caption{Perplexity on the language-specific subset of Wikipedia 50K articles. \textbf{Bold} and \underline{underlined} indicate the best and second-best perplexities across adapted models with the same base model for each language. Note that perplexities are not comparable between models with \colorbox{gray!25}{grey} and others due to their difference in vocabulary.}
    \label{tab:perplexity}
\end{table}

\subsection{Additional Downstream Results} \label{subsec:additional_result}
Table \ref{tab:main_result_full} shows the full results with standard deviations when prompted in English, and Table \ref{tab:main_result_target} shows the full results with standard deviations when prompted in a target language.

\paragraph{Poor Performance with Random Initialization in English Prompting}
Although models adapted with Random are competitive to other approaches using in-language prompts (Tables \ref{tab:main_result} and \ref{tab:main_result_target}), this is not the case when prompting in English (Table \ref{tab:main_result_full}).
We see a substantial drop in performance, especially in the Japanese, Arabic, and Swahili \textsc{span} tasks with BLOOM-1B.
For larger models, Random adversely affects not only \textsc{span} but also performance in \textsc{sum}.
For instance, we observe up to 53.0\% (German CVA TigerBot-7B) and 60.0\% (Japanese CVA TigerBot-7B) performance degradation in zero-shot \textsc{sum} and \textsc{span}, respectively.
Our hypothesis is that random initialization can severely impair the ability of the LLM to understand English prompts, which cannot be fully recovered with LAPT due to the exclusion of English data.

\setlength{\tabcolsep}{3pt}
\renewcommand*{\arraystretch}{1.0}
\begin{table*}[t]
\begin{center}
\small
\resizebox{\linewidth}{!}{%
\begin{tabular}{lllll|llll|llll|llll}
\textbf{Approach} & \multicolumn{4}{c|}{\textbf{German}}  & \multicolumn{4}{c|}{\textbf{Japanese}} & \multicolumn{4}{c|}{\textbf{Arabic}} & \multicolumn{4}{c}{\textbf{Swahili}}\\

 & \multicolumn{1}{c}{\scriptsize \textsc{nli}} & \multicolumn{1}{c}{\scriptsize \textsc{mc}} & \multicolumn{1}{c}{\scriptsize \textsc{sum}} & \multicolumn{1}{c|}{\scriptsize \textsc{span}} 
 & \multicolumn{1}{c}{\scriptsize \textsc{nli}} & \multicolumn{1}{c}{\scriptsize \textsc{mc}} & \multicolumn{1}{c}{\scriptsize \textsc{sum}} & \multicolumn{1}{c|}{\scriptsize \textsc{span}} 
 & \multicolumn{1}{c}{\scriptsize \textsc{nli}} & \multicolumn{1}{c}{\scriptsize \textsc{mc}} & \multicolumn{1}{c}{\scriptsize \textsc{sum}} & \multicolumn{1}{c|}{\scriptsize \textsc{span}} 
& \multicolumn{1}{c}{\scriptsize \textsc{nli}} & \multicolumn{1}{c}{\scriptsize \textsc{mc}} & \multicolumn{1}{c}{\scriptsize \textsc{sum}} & \multicolumn{1}{c}{\scriptsize \textsc{span}} \\

\hline
\multicolumn{7}{l}{\textbf{BLOOM-1B}} & \multicolumn{2}{c}{\underline{\textbf{Zero-shot}}}\\
\rowcolor{gray!25}
~~Source & .34\textsubscript{.00} & .20\textsubscript{.00} & 14.9\textsubscript{0.7} & .08\textsubscript{.00} & .17\textsubscript{.00} & .25\textsubscript{.00} & 3.7\textsubscript{0.1} & .23\textsubscript{.00} & .35\textsubscript{.00} & .19\textsubscript{.00} & 9.9\textsubscript{0.1} & .17\textsubscript{.01} & .34\textsubscript{.00} & .21\textsubscript{.00} & 10.6\textsubscript{0.4} & .08\textsubscript{.00}\\
\rowcolor{gray!25}
~~LAPT & .33\textsubscript{.00} & .21\textsubscript{.01} & 17.0\textsubscript{0.1} & .09\textsubscript{.01} & .17\textsubscript{.00} & .20\textsubscript{.00} & 17.2\textsubscript{0.3} & .25\textsubscript{.00} & .34\textsubscript{.00} & .18\textsubscript{.00} & 11.6\textsubscript{0.1} & .14\textsubscript{.01} & .34\textsubscript{.00} & .18\textsubscript{.00} & 9.7\textsubscript{0.1} & .08\textsubscript{.00}\\
\hline
\end{tabular}%
}
\caption{Mean performance over five runs with standard deviations when prompting in English on 500 randomly selected test samples for each dataset.
The baselines are in \colorbox{gray!25}{grey}.
\textbf{Bold} indicates comparable or better results than the baselines.
Darker \textcolor{blue}{blue} and  \textcolor{red}{red} shades indicate higher positive and negative relative performance change over Source per language and task, respectively.
}
\label{tab:main_result_full}
\end{center}
\end{table*}

\renewcommand*{\arraystretch}{1.0}
\begin{table*}[t]
\begin{center}
\small
\resizebox{\linewidth}{!}{%
\begin{tabular}{lllll|llll|llll|llll}
\textbf{Approach} & \multicolumn{4}{c|}{\textbf{German}}  & \multicolumn{4}{c|}{\textbf{Japanese}} & \multicolumn{4}{c|}{\textbf{Arabic}} & \multicolumn{4}{c}{\textbf{Swahili}}\\

 & \multicolumn{1}{c}{\scriptsize \textsc{nli}} & \multicolumn{1}{c}{\scriptsize \textsc{mc}} & \multicolumn{1}{c}{\scriptsize \textsc{sum}} & \multicolumn{1}{c|}{\scriptsize \textsc{span}} 
 & \multicolumn{1}{c}{\scriptsize \textsc{nli}} & \multicolumn{1}{c}{\scriptsize \textsc{mc}} & \multicolumn{1}{c}{\scriptsize \textsc{sum}} & \multicolumn{1}{c|}{\scriptsize \textsc{span}} 
 & \multicolumn{1}{c}{\scriptsize \textsc{nli}} & \multicolumn{1}{c}{\scriptsize \textsc{mc}} & \multicolumn{1}{c}{\scriptsize \textsc{sum}} & \multicolumn{1}{c|}{\scriptsize \textsc{span}} 
& \multicolumn{1}{c}{\scriptsize \textsc{nli}} & \multicolumn{1}{c}{\scriptsize \textsc{mc}} & \multicolumn{1}{c}{\scriptsize \textsc{sum}} & \multicolumn{1}{c}{\scriptsize \textsc{span}} \\

\hline
\multicolumn{7}{l}{\textbf{BLOOM-1B}} & \multicolumn{2}{c}{\underline{\textbf{Zero-shot}}}\\
\rowcolor{gray!25}
~~Source & .35\textsubscript{.00} & .21\textsubscript{.00} & 17.8\textsubscript{0.3} & .06\textsubscript{.00} & .29\textsubscript{.00} & .20\textsubscript{.00} & 18.2\textsubscript{0.3} & .22\textsubscript{.00} & .31\textsubscript{.00} & .20\textsubscript{.00} & 12.0\textsubscript{0.2} & .15\textsubscript{.01} & .35\textsubscript{.00} & .22\textsubscript{.00} & 12.0\textsubscript{0.3} & .03\textsubscript{.00}\\
\rowcolor{gray!25}
~~LAPT & .34\textsubscript{.01} & .22\textsubscript{.01} & 14.3\textsubscript{0.2} & .09\textsubscript{.00} & .28\textsubscript{.00} & .20\textsubscript{.00} & 20.7\textsubscript{0.2} & .26\textsubscript{.00} & .31\textsubscript{.00} & .19\textsubscript{.01} & 11.4\textsubscript{0.1} & .13\textsubscript{.01} & .35\textsubscript{.00} & .18\textsubscript{.00} & 7.7\textsubscript{0.1} & .07\textsubscript{.00}\\
\hline
\end{tabular}%
}
\caption{Mean performance over five runs with standard deviations when prompting in a target language on 500 randomly selected test samples for each dataset.
The baselines are in \colorbox{gray!25}{grey}.
\textbf{Bold} indicates comparable or better results than the baselines.
Darker \textcolor{blue}{blue} and  \textcolor{red}{red} shades indicate higher positive and negative relative performance change over Source per language and task, respectively.
}
\label{tab:main_result_target}
\end{center}
\end{table*}

\subsection{English Downstream Performance}
Table \ref{tab:eng_result} shows the results on the English datasets.
Despite the entire replacement of embeddings for CVA approaches, their adapted models exhibit comparable or better results in most of the tasks for BLOOM, except for \textsc{span}, where Source showed the best result followed by LAPT.
This can be ascribed to the following reasons:
First, LAPT can retain more source model knowledge than CVA approaches, as their embeddings have not changed.
Second, \textsc{span} can be seen as a challenging task as it requires more linguistic understanding of a prompt than simply classifying a text as in \textsc{nli} and \textsc{mc}.
We, therefore, see such a huge performance difference in \textsc{span} since CVA approaches lost more source linguistic knowledge than LAPT counterparts in exchange for faster inference in a target language.

For TigerBot and Mistral, which are not as multilingual as BLOOM, we see quite similar trends observed in \S\ref{subsec:performance} in that (1) models with CVA fail to achieve competitive downstream performance to the baselines and (2) their few-shot performance is far lower than LAPT.
These results suggest that there can be a relationship between downstream performance in a target language and those in English, and maintaining competitive downstream performance to the baselines in English might be a key to improving models with CVA in terms of their downstream performance.

\setlength{\tabcolsep}{3pt}
\renewcommand*{\arraystretch}{1.0}
\begin{table*}[t]
\begin{center}
\small
\resizebox{\linewidth}{!}{%
\begin{tabular}{lllll|llll|llll|llll}
\textbf{Approach} & \multicolumn{4}{c|}{\textbf{\textsc{nli}}}  & \multicolumn{4}{c|}{\textbf{\textsc{mc}}} & \multicolumn{4}{c|}{\textbf{\textsc{sum}}} & \multicolumn{4}{c}{\textbf{\textsc{span}}}\\

 & \multicolumn{1}{c}{\scriptsize de} & \multicolumn{1}{c}{\scriptsize ja} & \multicolumn{1}{c}{\scriptsize ar} & \multicolumn{1}{c|}{\scriptsize sw} 
 & \multicolumn{1}{c}{\scriptsize de} & \multicolumn{1}{c}{\scriptsize ja} & \multicolumn{1}{c}{\scriptsize ar} & \multicolumn{1}{c|}{\scriptsize sw} 
 & \multicolumn{1}{c}{\scriptsize de} & \multicolumn{1}{c}{\scriptsize ja} & \multicolumn{1}{c}{\scriptsize ar} & \multicolumn{1}{c|}{\scriptsize sw} 
& \multicolumn{1}{c}{\scriptsize de} & \multicolumn{1}{c}{\scriptsize ja} & \multicolumn{1}{c}{\scriptsize ar} & \multicolumn{1}{c}{\scriptsize sw}  \\

\hline
\multicolumn{7}{l}{\textbf{BLOOM-1B}} & \multicolumn{2}{c}{\underline{\textbf{Zero-shot}}}\\
\rowcolor{gray!25}
~~Source & \multicolumn{4}{c|}{.34\textsubscript{.00}} & \multicolumn{4}{c|}{.18\textsubscript{.00}} & \multicolumn{4}{c|}{11.2\textsubscript{0.1}} & \multicolumn{4}{c}{.21\textsubscript{.01}}\\
\rowcolor{gray!25}
~~LAPT & .35\textsubscript{.01} & .33\textsubscript{.00} & .33\textsubscript{.00} & .33\textsubscript{.00} & .21\textsubscript{.01} & .20\textsubscript{.00} & .17\textsubscript{.01} & .18\textsubscript{.01} & 9.7\textsubscript{0.1} & 10.4\textsubscript{0.0} & 10.6\textsubscript{0.1} & 10.1\textsubscript{0.0} & .15\textsubscript{.01} & .18\textsubscript{.00} & .17\textsubscript{.00} & .15\textsubscript{.01}\\
~~+ Heuristics & \cellcolor[rgb]{0.99,0.99,1.00}\textbf{.35}\textsubscript{.00} & \cellcolor[rgb]{1.00,1.00,1.00}\textbf{.34}\textsubscript{.00} & \cellcolor[rgb]{0.99,0.99,1.00}\textbf{.35}\textsubscript{.00} & \cellcolor[rgb]{0.97,0.97,1.00}\textbf{.36}\textsubscript{.00} & \cellcolor[rgb]{0.94,0.94,1.00}.20\textsubscript{.00} & \cellcolor[rgb]{0.94,0.94,1.00}\textbf{.20}\textsubscript{.00} & \cellcolor[rgb]{0.92,0.92,1.00}\textbf{.21}\textsubscript{.00} & \cellcolor[rgb]{0.94,0.94,1.00}\textbf{.20}\textsubscript{.00} & \cellcolor[rgb]{0.99,0.99,1.00}\textbf{11.2}\textsubscript{0.1} & \cellcolor[rgb]{0.97,0.97,1.00}\textbf{11.7}\textsubscript{0.1} & \cellcolor[rgb]{0.94,0.94,1.00}\textbf{12.3}\textsubscript{0.1} & \cellcolor[rgb]{1.00,0.96,0.96}10.1\textsubscript{0.2} & \cellcolor[rgb]{1.00,0.74,0.74}.10\textsubscript{.01} & \cellcolor[rgb]{1.00,0.64,0.64}.06\textsubscript{.01} & \cellcolor[rgb]{1.00,0.66,0.66}.07\textsubscript{.00} & \cellcolor[rgb]{1.00,0.71,0.71}.09\textsubscript{.01}\\
~~+ CLP+ & \cellcolor[rgb]{0.99,0.99,1.00}\textbf{.35}\textsubscript{.00} & \cellcolor[rgb]{1.00,1.00,1.00}\textbf{.34}\textsubscript{.00} & \cellcolor[rgb]{1.00,0.97,0.97}.32\textsubscript{.00} & \cellcolor[rgb]{0.95,0.95,1.00}\textbf{.37}\textsubscript{.01} & \cellcolor[rgb]{0.97,0.97,1.00}.19\textsubscript{.00} & \cellcolor[rgb]{0.94,0.94,1.00}\textbf{.20}\textsubscript{.00} & \cellcolor[rgb]{0.94,0.94,1.00}\textbf{.20}\textsubscript{.00} & \cellcolor[rgb]{0.94,0.94,1.00}\textbf{.20}\textsubscript{.00} & \cellcolor[rgb]{1.00,0.99,0.99}\textbf{10.8}\textsubscript{0.1} & \cellcolor[rgb]{0.95,0.95,1.00}\textbf{12.1}\textsubscript{0.1} & \cellcolor[rgb]{0.94,0.94,1.00}\textbf{12.4}\textsubscript{0.1} & \cellcolor[rgb]{1.00,0.97,0.97}10.4\textsubscript{0.2} & \cellcolor[rgb]{1.00,0.79,0.79}.12\textsubscript{.00} & \cellcolor[rgb]{1.00,0.76,0.76}.11\textsubscript{.00} & \cellcolor[rgb]{1.00,0.66,0.66}.07\textsubscript{.00} & \cellcolor[rgb]{1.00,0.69,0.69}.08\textsubscript{.00}\\

\hline
\multicolumn{6}{l}{\textbf{BLOOM-7B}}\\
\rowcolor{gray!25}
~~Source & \multicolumn{4}{c|}{.36\textsubscript{.00}} & \multicolumn{4}{c|}{.17\textsubscript{.00}} & \multicolumn{4}{c|}{11.1\textsubscript{0.1}} & \multicolumn{4}{c}{.31\textsubscript{.00}}\\
\rowcolor{gray!25}
~~LAPT & .34\textsubscript{.00} & .34\textsubscript{.00} & .34\textsubscript{.00} & .36\textsubscript{.00} & .20\textsubscript{.01} & .20\textsubscript{.01} & .20\textsubscript{.01} & .18\textsubscript{.01} & 10.8\textsubscript{0.0} & 11.0\textsubscript{0.0} & 10.9\textsubscript{0.0} & 10.6\textsubscript{0.1} & .25\textsubscript{.00} & .27\textsubscript{.01} & .28\textsubscript{.00} & .23\textsubscript{.00}\\
~~+ Heuristics & \cellcolor[rgb]{1.00,1.00,1.00}\textbf{.36}\textsubscript{.00} & \cellcolor[rgb]{1.00,0.97,0.97}.34\textsubscript{.00} & \cellcolor[rgb]{1.00,0.96,0.96}.33\textsubscript{.00} & \cellcolor[rgb]{1.00,1.00,1.00}\textbf{.36}\textsubscript{.00} & \cellcolor[rgb]{0.91,0.91,1.00}\textbf{.20}\textsubscript{.00} & \cellcolor[rgb]{0.91,0.91,1.00}\textbf{.20}\textsubscript{.00} & \cellcolor[rgb]{0.91,0.91,1.00}\textbf{.20}\textsubscript{.00} & \cellcolor[rgb]{0.94,0.94,1.00}\textbf{.19}\textsubscript{.00} & \cellcolor[rgb]{0.99,0.99,1.00}\textbf{11.2}\textsubscript{0.1} & \cellcolor[rgb]{0.99,0.99,1.00}\textbf{11.1}\textsubscript{0.1} & \cellcolor[rgb]{0.95,0.95,1.00}\textbf{12.2}\textsubscript{0.1} & \cellcolor[rgb]{1.00,0.98,0.98}10.5\textsubscript{0.0} & \cellcolor[rgb]{1.00,0.89,0.89}.24\textsubscript{.01} & \cellcolor[rgb]{1.00,0.81,0.81}.19\textsubscript{.00} & \cellcolor[rgb]{1.00,0.77,0.77}.17\textsubscript{.00} & \cellcolor[rgb]{1.00,0.77,0.77}.17\textsubscript{.00}\\
~~+ CLP+ & \cellcolor[rgb]{1.00,1.00,1.00}\textbf{.36}\textsubscript{.00} & \cellcolor[rgb]{1.00,0.97,0.97}.34\textsubscript{.00} & \cellcolor[rgb]{1.00,0.96,0.96}.33\textsubscript{.00} & \cellcolor[rgb]{0.97,0.97,1.00}\textbf{.38}\textsubscript{.00} & \cellcolor[rgb]{0.91,0.91,1.00}\textbf{.20}\textsubscript{.00} & \cellcolor[rgb]{0.91,0.91,1.00}\textbf{.20}\textsubscript{.00} & \cellcolor[rgb]{0.91,0.91,1.00}\textbf{.20}\textsubscript{.00} & \cellcolor[rgb]{0.91,0.91,1.00}\textbf{.20}\textsubscript{.00} & \cellcolor[rgb]{1.00,0.98,0.98}10.5\textsubscript{0.1} & \cellcolor[rgb]{0.96,0.96,1.00}\textbf{11.8}\textsubscript{0.1} & \cellcolor[rgb]{0.94,0.94,1.00}\textbf{12.4}\textsubscript{0.0} & \cellcolor[rgb]{0.99,0.99,1.00}\textbf{11.1}\textsubscript{0.1} & \cellcolor[rgb]{1.00,0.84,0.84}.21\textsubscript{.00} & \cellcolor[rgb]{1.00,0.66,0.66}.10\textsubscript{.01} & \cellcolor[rgb]{1.00,0.85,0.85}.22\textsubscript{.00} & \cellcolor[rgb]{1.00,0.81,0.81}.19\textsubscript{.00}\\

\hline
\multicolumn{6}{l}{\textbf{TigerBot-7B}}\\
\rowcolor{gray!25}
~~Source & \multicolumn{4}{c|}{.48\textsubscript{.00}} & \multicolumn{4}{c|}{.29\textsubscript{.00}} & \multicolumn{4}{c|}{12.7\textsubscript{0.1}} & \multicolumn{4}{c}{.42\textsubscript{.01}}\\
\rowcolor{gray!25}
~~LAPT & .39\textsubscript{.00} & .49\textsubscript{.01} & .45\textsubscript{.01} & .45\textsubscript{.01} & .23\textsubscript{.00} & .25\textsubscript{.00} & .25\textsubscript{.00} & .28\textsubscript{.01} & 11.6\textsubscript{0.1} & 11.9\textsubscript{0.1} & 12.2\textsubscript{0.1} & 11.0\textsubscript{0.1} & .31\textsubscript{.00} & .34\textsubscript{.01} & .27\textsubscript{.00} & .35\textsubscript{.01}\\
~~+ Heuristics & \cellcolor[rgb]{1.00,0.88,0.88}.37\textsubscript{.00} & \cellcolor[rgb]{1.00,0.85,0.85}.34\textsubscript{.00} & \cellcolor[rgb]{1.00,0.86,0.86}.35\textsubscript{.00} & \cellcolor[rgb]{1.00,0.82,0.82}.31\textsubscript{.00} & \cellcolor[rgb]{1.00,0.86,0.86}.21\textsubscript{.00} & \cellcolor[rgb]{1.00,0.91,0.91}.24\textsubscript{.00} & \cellcolor[rgb]{1.00,0.85,0.85}.20\textsubscript{.00} & \cellcolor[rgb]{1.00,0.85,0.85}.20\textsubscript{.00} & \cellcolor[rgb]{1.00,0.89,0.89}10.2\textsubscript{0.1} & \cellcolor[rgb]{1.00,0.97,0.97}12.1\textsubscript{0.1} & \cellcolor[rgb]{1.00,0.88,0.88}9.8\textsubscript{0.1} & \cellcolor[rgb]{1.00,0.68,0.68}4.8\textsubscript{0.1} & \cellcolor[rgb]{1.00,0.68,0.68}.15\textsubscript{.00} & \cellcolor[rgb]{1.00,0.79,0.79}.24\textsubscript{.01} & \cellcolor[rgb]{1.00,0.54,0.54}.03\textsubscript{.00} & \cellcolor[rgb]{1.00,0.52,0.52}.02\textsubscript{.00}\\
~~+ CLP+ & \cellcolor[rgb]{1.00,0.91,0.91}.39\textsubscript{.00} & \cellcolor[rgb]{1.00,0.88,0.88}.36\textsubscript{.00} & \cellcolor[rgb]{1.00,0.88,0.88}.36\textsubscript{.00} & \cellcolor[rgb]{1.00,0.82,0.82}.31\textsubscript{.00} & \cellcolor[rgb]{1.00,0.91,0.91}.24\textsubscript{.00} & \cellcolor[rgb]{1.00,0.91,0.91}.24\textsubscript{.00} & \cellcolor[rgb]{1.00,0.86,0.86}.21\textsubscript{.00} & \cellcolor[rgb]{1.00,0.85,0.85}.20\textsubscript{.00} & \cellcolor[rgb]{1.00,0.90,0.90}10.4\textsubscript{0.1} & \cellcolor[rgb]{1.00,0.98,0.98}12.5\textsubscript{0.1} & \cellcolor[rgb]{1.00,0.94,0.94}11.3\textsubscript{0.1} & \cellcolor[rgb]{1.00,0.74,0.74}6.3\textsubscript{0.1} & \cellcolor[rgb]{1.00,0.74,0.74}.20\textsubscript{.00} & \cellcolor[rgb]{1.00,0.86,0.86}.30\textsubscript{.00} & \cellcolor[rgb]{1.00,0.74,0.74}.20\textsubscript{.00} & \cellcolor[rgb]{1.00,0.54,0.54}.03\textsubscript{.00}\\

\hline
\multicolumn{6}{l}{\textbf{Mistral-7B}}\\
\rowcolor{gray!25}
~~Source & \multicolumn{4}{c|}{.42\textsubscript{.00}} & \multicolumn{4}{c|}{.46\textsubscript{.00}} & \multicolumn{4}{c|}{12.4\textsubscript{0.2}} & \multicolumn{4}{c}{.44\textsubscript{.00}}\\
\rowcolor{gray!25}
~~LAPT & .36\textsubscript{.01} & .49\textsubscript{.01} & .45\textsubscript{.01} & .42\textsubscript{.00} & .34\textsubscript{.01} & .32\textsubscript{.01} & .28\textsubscript{.01} & .38\textsubscript{.01} & 11.6\textsubscript{0.0} & 11.3\textsubscript{0.1} & 8.1\textsubscript{0.2} & 10.6\textsubscript{0.1} & .39\textsubscript{.00} & .40\textsubscript{.01} & .28\textsubscript{.00} & .36\textsubscript{.00}\\
~~+ Heuristics & \cellcolor[rgb]{1.00,0.89,0.89}.33\textsubscript{.00} & \cellcolor[rgb]{1.00,0.96,0.96}.39\textsubscript{.00} & \cellcolor[rgb]{1.00,0.93,0.93}.36\textsubscript{.00} & \cellcolor[rgb]{1.00,0.89,0.89}.33\textsubscript{.00} & \cellcolor[rgb]{1.00,0.73,0.73}.21\textsubscript{.00} & \cellcolor[rgb]{1.00,0.73,0.73}.21\textsubscript{.00} & \cellcolor[rgb]{1.00,0.72,0.72}.20\textsubscript{.00} & \cellcolor[rgb]{1.00,0.72,0.72}.20\textsubscript{.00} & \cellcolor[rgb]{0.98,0.98,1.00}\textbf{12.5}\textsubscript{0.1} & \cellcolor[rgb]{0.96,0.96,1.00}\textbf{12.9}\textsubscript{0.1} & \cellcolor[rgb]{1.00,0.98,0.98}11.5\textsubscript{0.2} & \cellcolor[rgb]{1.00,0.85,0.85}8.5\textsubscript{0.2} & \cellcolor[rgb]{1.00,0.76,0.76}.23\textsubscript{.00} & \cellcolor[rgb]{1.00,0.84,0.84}.30\textsubscript{.00} & \cellcolor[rgb]{1.00,0.72,0.72}.19\textsubscript{.00} & \cellcolor[rgb]{1.00,0.57,0.57}.06\textsubscript{.00}\\
~~+ CLP+ & \cellcolor[rgb]{1.00,0.93,0.93}.36\textsubscript{.00} & \cellcolor[rgb]{1.00,0.94,0.94}.37\textsubscript{.00} & \cellcolor[rgb]{1.00,0.90,0.90}.34\textsubscript{.00} & \cellcolor[rgb]{1.00,0.87,0.87}.31\textsubscript{.01} & \cellcolor[rgb]{1.00,0.73,0.73}.21\textsubscript{.00} & \cellcolor[rgb]{1.00,0.77,0.77}.25\textsubscript{.00} & \cellcolor[rgb]{1.00,0.70,0.70}.19\textsubscript{.00} & \cellcolor[rgb]{1.00,0.74,0.74}.22\textsubscript{.00} & \cellcolor[rgb]{0.99,0.99,1.00}12.2\textsubscript{0.1} & \cellcolor[rgb]{0.95,0.95,1.00}\textbf{13.1}\textsubscript{0.1} & \cellcolor[rgb]{0.96,0.96,1.00}\textbf{12.9}\textsubscript{0.1} & \cellcolor[rgb]{1.00,0.94,0.94}10.6\textsubscript{0.1} & \cellcolor[rgb]{1.00,0.79,0.79}.26\textsubscript{.00} & \cellcolor[rgb]{1.00,0.81,0.81}.27\textsubscript{.01} & \cellcolor[rgb]{1.00,0.85,0.85}.31\textsubscript{.00} & \cellcolor[rgb]{1.00,0.70,0.70}.18\textsubscript{.00}\\

\hline\hline

\multicolumn{7}{l}{\textbf{BLOOM-1B}} & \multicolumn{2}{c}{\underline{\textbf{Few-shot}}}\\
\rowcolor{gray!25}
~~Source &\multicolumn{4}{c|}{.33\textsubscript{.00}} & \multicolumn{4}{c|}{.20\textsubscript{.00}} & \multicolumn{4}{c|}{-}  & \multicolumn{4}{c}{.28\textsubscript{.00}}\\
\rowcolor{gray!25}
~~LAPT & .34\textsubscript{.01} & .31\textsubscript{.00} & .32\textsubscript{.01} & .33\textsubscript{.00} & .17\textsubscript{.01} & .18\textsubscript{.01} & .19\textsubscript{.01} & .20\textsubscript{.01} & - & - & - & - & .20\textsubscript{.01} & .23\textsubscript{.01} & .25\textsubscript{.00} & .20\textsubscript{.01}\\
~~+ Heuristics & \cellcolor[rgb]{1.00,1.00,1.00}.33\textsubscript{.00} & \cellcolor[rgb]{0.99,0.99,1.00}\textbf{.34}\textsubscript{.00} & \cellcolor[rgb]{1.00,0.99,0.99}.32\textsubscript{.00} & \cellcolor[rgb]{0.99,0.99,1.00}\textbf{.34}\textsubscript{.00} & \cellcolor[rgb]{1.00,0.97,0.97}.19\textsubscript{.00} & \cellcolor[rgb]{1.00,1.00,1.00}\textbf{.20}\textsubscript{.00} & \cellcolor[rgb]{0.97,0.97,1.00}\textbf{.21}\textsubscript{.00} & \cellcolor[rgb]{1.00,0.97,0.97}.19\textsubscript{.00} & - & - & - & - & \cellcolor[rgb]{1.00,0.79,0.79}.16\textsubscript{.00} & \cellcolor[rgb]{1.00,0.70,0.70}.11\textsubscript{.00} & \cellcolor[rgb]{1.00,0.75,0.75}.14\textsubscript{.00} & \cellcolor[rgb]{1.00,0.80,0.80}.17\textsubscript{.01}\\
~~+ CLP+ & \cellcolor[rgb]{0.94,0.94,1.00}\textbf{.37}\textsubscript{.00} & \cellcolor[rgb]{0.99,0.99,1.00}\textbf{.34}\textsubscript{.00} & \cellcolor[rgb]{0.97,0.97,1.00}\textbf{.35}\textsubscript{.00} & \cellcolor[rgb]{0.97,0.97,1.00}\textbf{.35}\textsubscript{.00} & \cellcolor[rgb]{1.00,0.97,0.97}.19\textsubscript{.00} & \cellcolor[rgb]{0.97,0.97,1.00}\textbf{.21}\textsubscript{.00} & \cellcolor[rgb]{0.95,0.95,1.00}\textbf{.22}\textsubscript{.00} & \cellcolor[rgb]{1.00,1.00,1.00}\textbf{.20}\textsubscript{.00} & - & - & - & - & \cellcolor[rgb]{1.00,0.80,0.80}.17\textsubscript{.00} & \cellcolor[rgb]{1.00,0.68,0.68}.10\textsubscript{.01} & \cellcolor[rgb]{1.00,0.73,0.73}.13\textsubscript{.00} & \cellcolor[rgb]{1.00,0.82,0.82}.18\textsubscript{.00}\\

\hline
\multicolumn{6}{l}{\textbf{BLOOM-7B}}\\
\rowcolor{gray!25}
~~Source & \multicolumn{4}{c|}{.43\textsubscript{.00}} &  \multicolumn{4}{c|}{.21\textsubscript{.00}} &  \multicolumn{4}{c|}{-}  & \multicolumn{4}{c}{.39\textsubscript{.00}}\\
\rowcolor{gray!25}
~~LAPT & .36\textsubscript{.01} & .38\textsubscript{.01} & .40\textsubscript{.00} & .39\textsubscript{.01} & .20\textsubscript{.01} & .21\textsubscript{.01} & .21\textsubscript{.00} & .19\textsubscript{.00} & - & - & - & - & .36\textsubscript{.00} & .38\textsubscript{.00} & .38\textsubscript{.00} & .37\textsubscript{.00}\\
~~+ Heuristics & \cellcolor[rgb]{1.00,0.92,0.92}.36\textsubscript{.00} & \cellcolor[rgb]{1.00,0.91,0.91}.35\textsubscript{.00} & \cellcolor[rgb]{1.00,0.86,0.86}.31\textsubscript{.00} & \cellcolor[rgb]{1.00,0.87,0.87}.32\textsubscript{.00} & \cellcolor[rgb]{1.00,0.97,0.97}.20\textsubscript{.00} & \cellcolor[rgb]{1.00,0.95,0.95}.19\textsubscript{.00} & \cellcolor[rgb]{0.97,0.97,1.00}\textbf{.22}\textsubscript{.00} & \cellcolor[rgb]{1.00,1.00,1.00}\textbf{.21}\textsubscript{.00} & - & - & - & - & \cellcolor[rgb]{1.00,0.97,0.97}.37\textsubscript{.00} & \cellcolor[rgb]{1.00,0.92,0.92}.33\textsubscript{.00} & \cellcolor[rgb]{1.00,0.96,0.96}.36\textsubscript{.00} & \cellcolor[rgb]{1.00,0.94,0.94}.34\textsubscript{.00}\\
~~+ CLP+ & \cellcolor[rgb]{1.00,0.92,0.92}.36\textsubscript{.00} & \cellcolor[rgb]{1.00,0.88,0.88}.33\textsubscript{.00} & \cellcolor[rgb]{1.00,0.88,0.88}.33\textsubscript{.00} & \cellcolor[rgb]{1.00,0.90,0.90}.34\textsubscript{.00} & \cellcolor[rgb]{1.00,0.93,0.93}.18\textsubscript{.00} & \cellcolor[rgb]{1.00,1.00,1.00}\textbf{.21}\textsubscript{.00} & \cellcolor[rgb]{1.00,0.97,0.97}.20\textsubscript{.00} & \cellcolor[rgb]{1.00,0.97,0.97}.20\textsubscript{.00} & - & - & - & - & \cellcolor[rgb]{1.00,0.95,0.95}.35\textsubscript{.00} & \cellcolor[rgb]{1.00,0.94,0.94}.34\textsubscript{.00} & \cellcolor[rgb]{1.00,0.96,0.96}.36\textsubscript{.03} & \cellcolor[rgb]{1.00,0.96,0.96}.36\textsubscript{.00}\\

\hline
\multicolumn{6}{l}{\textbf{TigerBot-7B}}\\
\rowcolor{gray!25}
~~Source & \multicolumn{4}{c|}{.49\textsubscript{.00}} &  \multicolumn{4}{c|}{.58\textsubscript{.00}} & \multicolumn{4}{c|}{-}  & \multicolumn{4}{c}{.47\textsubscript{.00}}\\
\rowcolor{gray!25}
~~LAPT & .47\textsubscript{.01} & .56\textsubscript{.01} & .45\textsubscript{.00} & .56\textsubscript{.01} & .57\textsubscript{.00} & .58\textsubscript{.00} & .52\textsubscript{.00} & .52\textsubscript{.01} & - & - & - & - & .47\textsubscript{.00} & .46\textsubscript{.00} & .44\textsubscript{.00} & .47\textsubscript{.00}\\
~~+ Heuristics & \cellcolor[rgb]{1.00,0.87,0.87}.36\textsubscript{.00} & \cellcolor[rgb]{1.00,0.86,0.86}.35\textsubscript{.00} & \cellcolor[rgb]{1.00,0.88,0.88}.37\textsubscript{.00} & \cellcolor[rgb]{1.00,0.85,0.85}.34\textsubscript{.00} & \cellcolor[rgb]{1.00,0.67,0.67}.20\textsubscript{.00} & \cellcolor[rgb]{1.00,0.77,0.77}.31\textsubscript{.00} & \cellcolor[rgb]{1.00,0.70,0.70}.23\textsubscript{.00} & \cellcolor[rgb]{1.00,0.66,0.66}.19\textsubscript{.00} & - & - & - & - & \cellcolor[rgb]{1.00,0.75,0.75}.24\textsubscript{.00} & \cellcolor[rgb]{1.00,0.94,0.94}.41\textsubscript{.01} & \cellcolor[rgb]{1.00,0.53,0.53}.03\textsubscript{.00} & \cellcolor[rgb]{1.00,0.51,0.51}.01\textsubscript{.00}\\
~~+ CLP+ & \cellcolor[rgb]{1.00,0.93,0.93}.42\textsubscript{.00} & \cellcolor[rgb]{1.00,0.85,0.85}.34\textsubscript{.00} & \cellcolor[rgb]{1.00,0.88,0.88}.37\textsubscript{.00} & \cellcolor[rgb]{1.00,0.87,0.87}.36\textsubscript{.00} & \cellcolor[rgb]{1.00,0.77,0.77}.32\textsubscript{.00} & \cellcolor[rgb]{1.00,0.70,0.70}.24\textsubscript{.00} & \cellcolor[rgb]{1.00,0.66,0.66}.19\textsubscript{.00} & \cellcolor[rgb]{1.00,0.66,0.66}.19\textsubscript{.00} & - & - & - & - & \cellcolor[rgb]{1.00,0.89,0.89}.37\textsubscript{.00} & \cellcolor[rgb]{1.00,0.96,0.96}.43\textsubscript{.00} & \cellcolor[rgb]{1.00,0.81,0.81}.29\textsubscript{.00} & \cellcolor[rgb]{1.00,0.52,0.52}.02\textsubscript{.00}\\
\hline
\multicolumn{6}{l}{\textbf{Mistral-7B}}\\
\rowcolor{gray!25}
~~Source & \multicolumn{4}{c|}{.60\textsubscript{.00}} &  \multicolumn{4}{c|}{.66\textsubscript{.00}} & \multicolumn{4}{c|}{-} & \multicolumn{4}{c}{.51\textsubscript{.00}}\\
\rowcolor{gray!25}
~~LAPT & .55\textsubscript{.00} & .53\textsubscript{.01} & .49\textsubscript{.01} & .56\textsubscript{.01} & .62\textsubscript{.00} & .59\textsubscript{.01} & .57\textsubscript{.01} & .63\textsubscript{.01} & - & - & - & - & .38\textsubscript{.00} & .49\textsubscript{.00} & .31\textsubscript{.00} & .49\textsubscript{.00}\\
~~+ Heuristics & \cellcolor[rgb]{1.00,0.86,0.86}.43\textsubscript{.00} & \cellcolor[rgb]{1.00,0.79,0.79}.35\textsubscript{.00} & \cellcolor[rgb]{1.00,0.80,0.80}.36\textsubscript{.00} & \cellcolor[rgb]{1.00,0.77,0.77}.33\textsubscript{.00} & \cellcolor[rgb]{1.00,0.75,0.75}.33\textsubscript{.00} & \cellcolor[rgb]{1.00,0.74,0.74}.31\textsubscript{.00} & \cellcolor[rgb]{1.00,0.65,0.65}.20\textsubscript{.00} & \cellcolor[rgb]{1.00,0.64,0.64}.19\textsubscript{.00} & - & - & - & - & \cellcolor[rgb]{1.00,0.76,0.76}.27\textsubscript{.00} & \cellcolor[rgb]{1.00,0.96,0.96}.47\textsubscript{.00} & \cellcolor[rgb]{1.00,0.81,0.81}.32\textsubscript{.00} & \cellcolor[rgb]{1.00,0.55,0.55}.05\textsubscript{.00}\\
~~+ CLP+ & \cellcolor[rgb]{1.00,0.82,0.82}.38\textsubscript{.00} & \cellcolor[rgb]{1.00,0.78,0.78}.34\textsubscript{.00} & \cellcolor[rgb]{1.00,0.82,0.82}.38\textsubscript{.00} & \cellcolor[rgb]{1.00,0.81,0.81}.37\textsubscript{.00} & \cellcolor[rgb]{1.00,0.84,0.84}.45\textsubscript{.00} & \cellcolor[rgb]{1.00,0.74,0.74}.31\textsubscript{.00} & \cellcolor[rgb]{1.00,0.70,0.70}.26\textsubscript{.00} & \cellcolor[rgb]{1.00,0.66,0.66}.21\textsubscript{.00} & - & - & - & - & \cellcolor[rgb]{1.00,0.83,0.83}.34\textsubscript{.00} & \cellcolor[rgb]{1.00,0.94,0.94}.45\textsubscript{.00} & \cellcolor[rgb]{1.00,0.93,0.93}.44\textsubscript{.00} & \cellcolor[rgb]{1.00,0.77,0.77}.28\textsubscript{.01}\\
\hline
\end{tabular}%
}
\caption{Mean performance over five runs with standard deviations on 500 randomly selected test samples for each English dataset.
The baselines are in \colorbox{gray!25}{grey}.
\textbf{Bold} indicates comparable or better results than the baselines.
Darker \textcolor{blue}{blue} and  \textcolor{red}{red} shades indicate higher positive and negative relative performance change over Source per language and task, respectively.
}
\label{tab:eng_result}
\end{center}
\end{table*}

\subsection{How helpful is LAPT for LLMs with cross-lingual vocabulary adaptation?}

\begin{figure}[t!]
\begin{center}
\includegraphics[width=0.95\columnwidth]{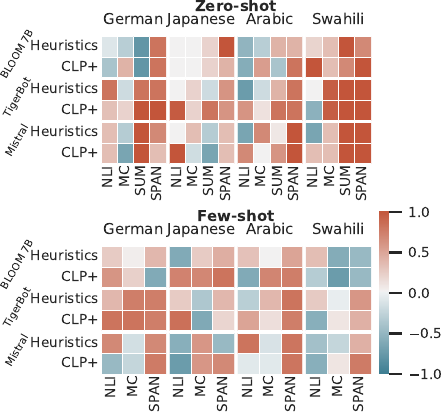}
\caption{
Kendall's $\tau$ correlation between the number of LAPT steps and performance (English prompting).
}
\label{fig:step_ablation_target}
\end{center}
\end{figure}

Figure \ref{fig:step_ablation_target} visualizes Kendall's tau correlation coefficient between the number of LAPT steps and downstream performance when prompted in English.
Similar to Figure \ref{fig:step_ablation}, we observe that LAPT helped improve downstream performance in both zero-shot and few-shot settings even when prompted in a target language in 65.6\% and 63.5\% of the cases, respectively.

\subsection{Loss Curves}
\begin{figure*}[t]
\centering
\includegraphics[width=\textwidth]{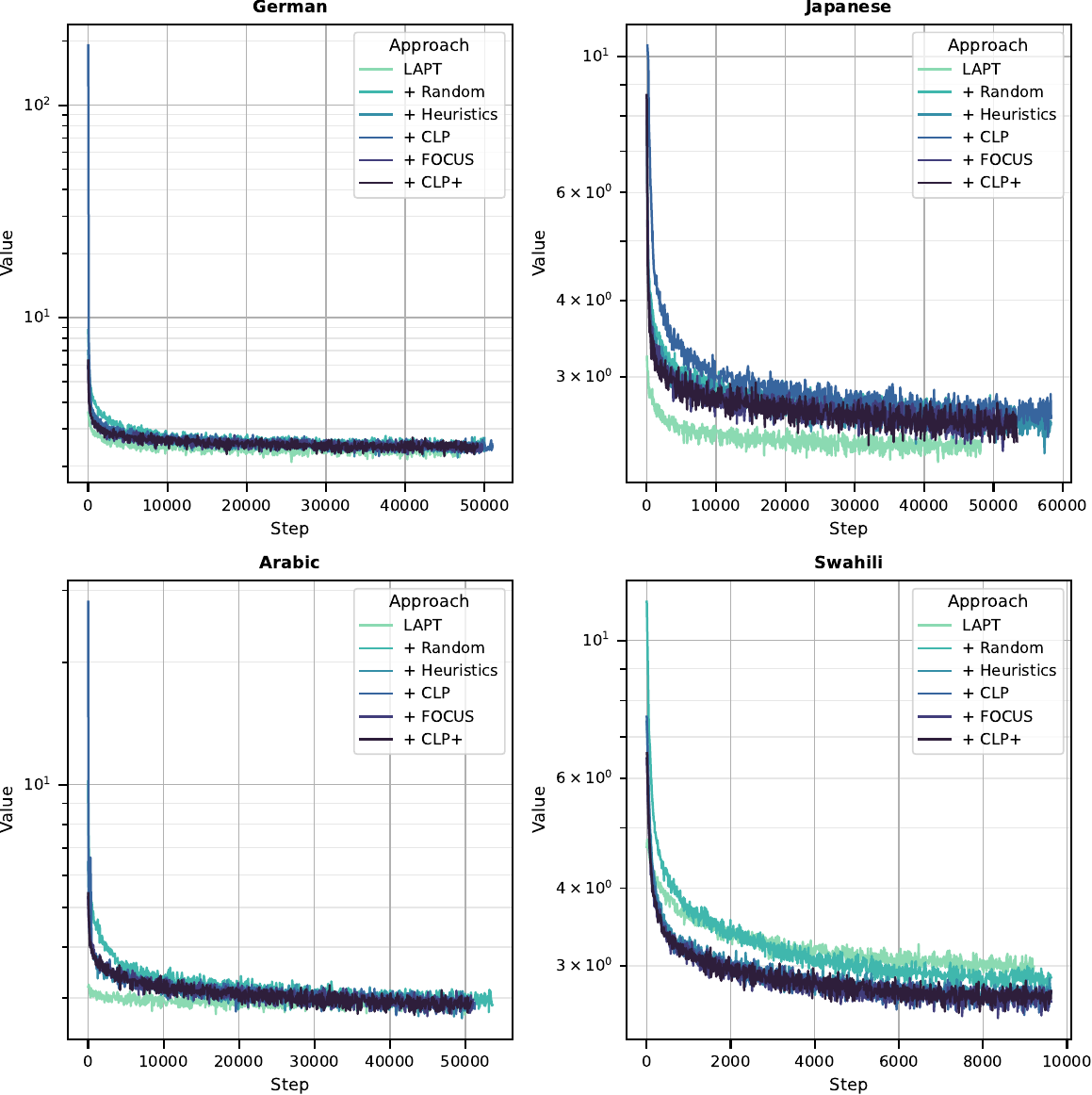}
\caption{LAPT loss curves for BLOOM-1B}
\label{figure:loss_1b1}
\end{figure*}

\begin{figure*}[t]
\centering
\includegraphics[width=\textwidth]{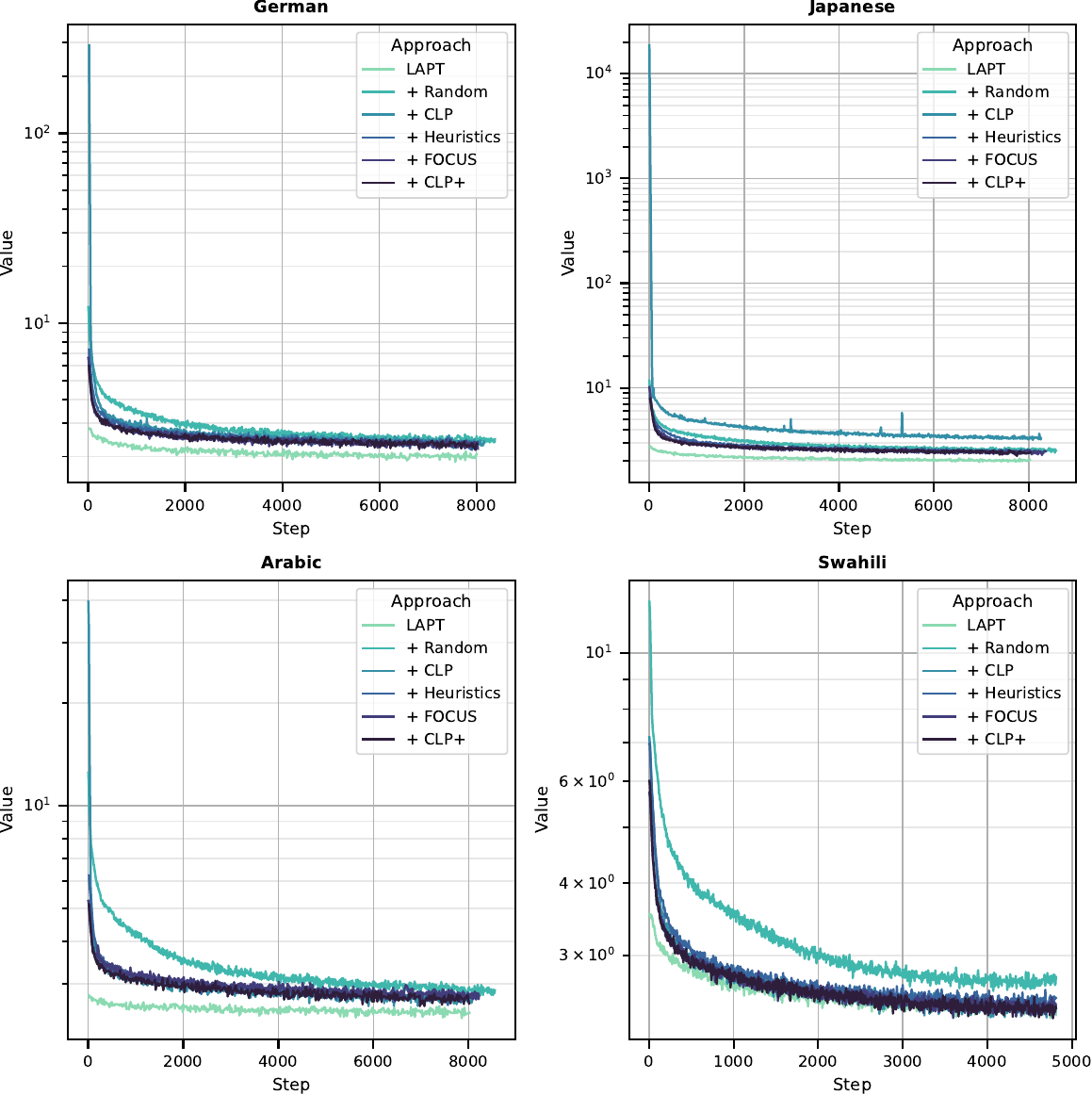}
\caption{LAPT loss curves for BLOOM-7B}
\label{figure:loss_7b1}
\end{figure*}

\begin{figure*}[t]
\centering
\includegraphics[width=\textwidth]{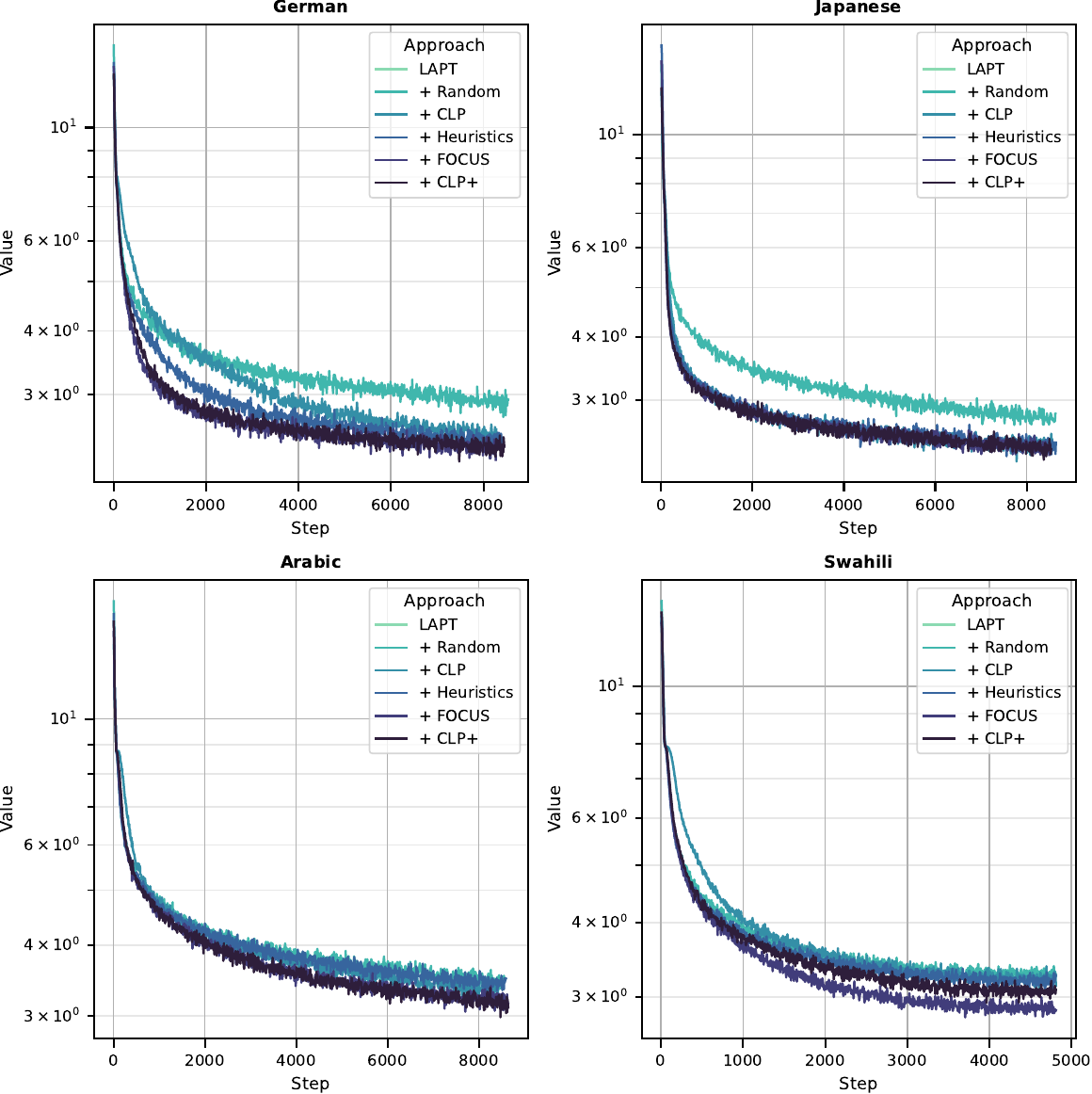}
\caption{LAPT loss curves for TigerBot-7B}
\label{figure:loss_tigerbot}
\end{figure*}

\begin{figure*}[t]
\centering
\includegraphics[width=\textwidth]{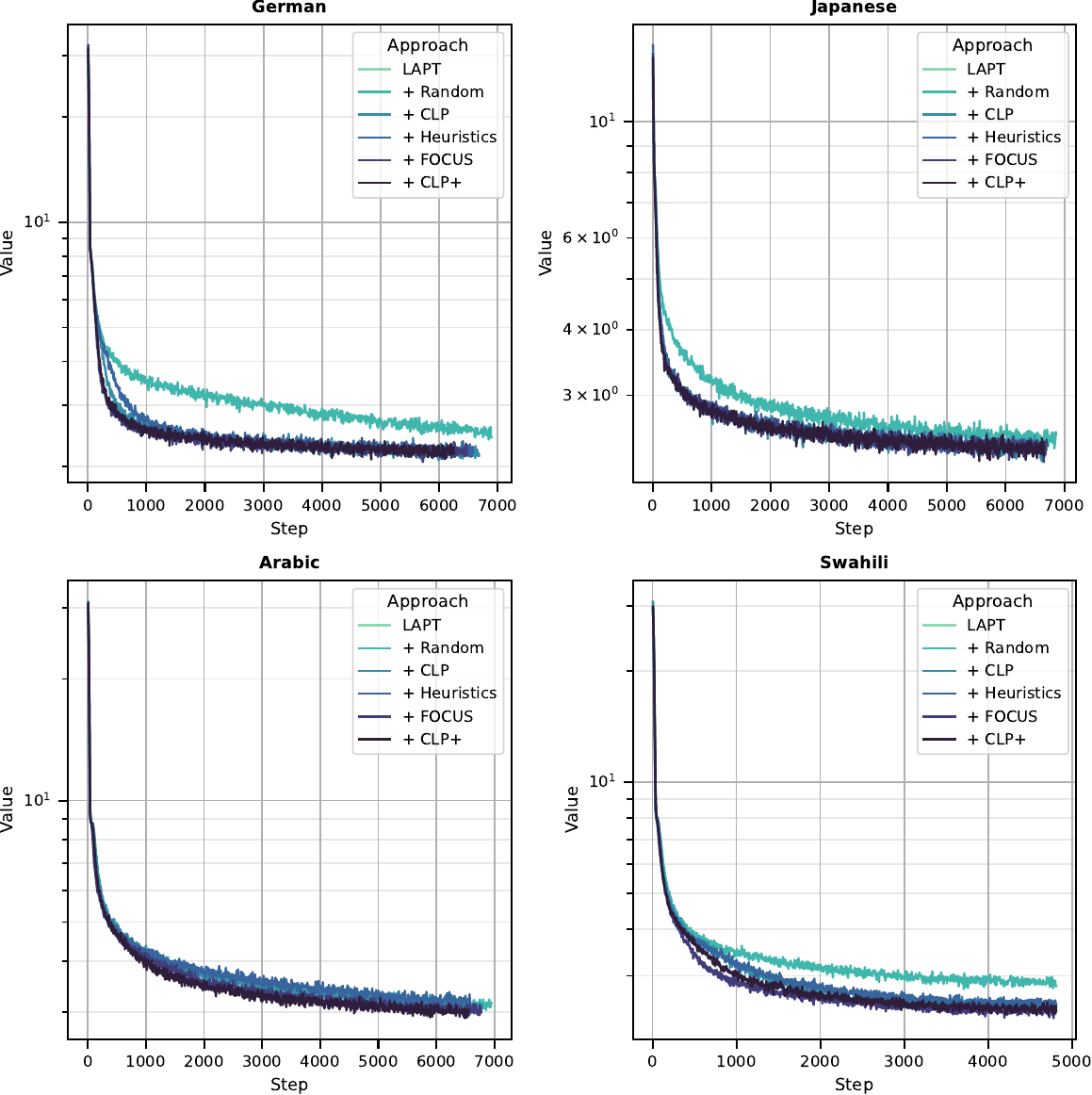}
\caption{LAPT loss curves for Mistral-7B}
\label{figure:loss_mistral}
\end{figure*}

Figures \ref{figure:loss_1b1} to \ref{figure:loss_mistral} show the loss curves in LAPT for each model setting.

\subsection{Kendall's Tau Correlation Coefficient for Figure \ref{fig:rank_ablation}}

Table \ref{tab:kendall_lora} lists all Kendall's tau correlation coefficients corresponding to Figure \ref{fig:rank_ablation} in \S\ref{subsec:rank}.
Models using CVA do not exhibit a strong correlation in the zero-shot setting, ranging from -0.226 to 0.173.
The only exception is Heuristics in Japanese with English prompting (0.45).
We observe a positive correlation ranging (0.59-0.889) in the few-shot setting, except for CLP+ in Japanese with English prompting (-0.95).

\begin{table}[H]
\begin{center}
\small
\begin{tabular}{lllll}
\textbf{Approach} & \multicolumn{2}{c}{\textbf{Japanese}} & \multicolumn{2}{c}{\textbf{Swahili}} \\
 & \multicolumn{1}{c}{\scriptsize English} & \multicolumn{1}{c}{\scriptsize Target} & \multicolumn{1}{c}{\scriptsize English} & \multicolumn{1}{c}{\scriptsize Target} \\ 
 \hline
 \multicolumn{5}{c}{\underline{\textbf{Zero-shot}}}\\
LAPT & 0.720 & 0.333 & 0.788 & 0.187 \\
+ Heuristics & 0.453 & 0.173 & 0.173 & 0.160 \\
+ CLP+ & -0.160 & -0.106 & -0.226 & 0.066 \\ 
\hline
 \multicolumn{5}{c}{\underline{\textbf{Few-shot}}}\\
LAPT & 0.626 & 1.00 & 0.453 & 0.906 \\
+ Heuristics & 0.706 & 0.600 & 0.591 & 0.701 \\
+ CLP+ & -0.946 & 0.626 & 0.886 & 0.756 \\ 
\hline
\end{tabular}
\caption{Kendall's tau correlation coefficients corresponding to Figure \ref{fig:rank_ablation}. We include LAPT results for reference.}
\label{tab:kendall_lora}
\end{center}
\end{table}

\clearpage

\end{document}